\documentclass[a4paper, conference, nofonttune]{IEEEtran}

\usepackage[english]{babel}
\usepackage[utf8]{inputenc}
\usepackage[T1]{fontenc}

\usepackage{pgf}
\usepackage{tikz}
\usepackage{pgfplots}
\usetikzlibrary{arrows,shapes,shapes.geometric,automata,petri,backgrounds,patterns}
\usepgfplotslibrary{fillbetween}
\pgfplotsset{compat=1.13}

\usepackage{tabu}
\usepackage{booktabs}
\usepackage{multirow}
\usepackage{colortbl}

\usepackage{mathtools}
\usepackage{amsmath}
\usepackage{amssymb}
\usepackage{amsthm}
\theoremstyle{definition}
\newtheorem{definition}{Definition}

\usepackage{fmtcount}
\usepackage[group-digits=integer, output-decimal-marker={.}, binary-units=true]{siunitx}
\usepackage{algorithm}
\usepackage{algpseudocode}
\makeatletter
\algrenewcommand{\algorithmiccomment}[1]{\textit{\texttt{\scriptsize //#1}}}
\algrenewcommand\algorithmicindent{1em}
\algrenewcommand\ALG@beginalgorithmic{\small}
\makeatother

\usepackage[caption=false,font=footnotesize]{subfig}

\usepackage{cite}

\usepackage[inline]{enumitem}

\usepackage{microtype}

\usepackage{todonotes}

\usepackage{varwidth}

\newcommand{\algorithmname}{Algorithm}
\newcommand{\sectionname}{Section}
\newcommand{\definitionname}{Def.}
\newcommand{\proposedname}{\textit{C2D2}}

\begin{document}
\bstctlcite{BIBTEX:IEEEConfig}

\title{A Conformance Checking-based Approach\\for Drift Detection in Business Processes}

\author{
  Víctor Gallego-Fontenla, Juan C. Vidal, and Manuel Lama\\
  Centro Singular de Investigación en Tecnoloxías Intelixentes (CiTIUS),\\
  Universidade de Santiago de Compostela, Galicia, Spain.\\
  \mbox{E-mail: \{victorjose.gallego, juan.vidal, manuel.lama\}@usc.es}\\
  \\
}

\maketitle

\begin{abstract}\label{section:abstract}
  Real life business processes change over time, in both planned and unexpected ways.
  The detection of these changes is crucial for organizations to ensure that the expected and the real behavior are as similar as possible.
  These changes over time are called concept drift and its detection is a big challenge in process mining since the inherent complexity of the data makes difficult distinguishing between a change and an anomalous execution.
  In this paper, we present \proposedname~(Conformance Checking-based Drift Detection), a new approach to detect sudden control-flow changes in the process models from event traces.
  \proposedname~combines discovery techniques with conformance checking methods to perform an offline detection.
  Our approach has been validated with a synthetic benchmarking dataset formed by 68 logs, showing an improvement in the accuracy ($F_{score}$) while maintaining a minimum delay in the drift detection.
\end{abstract}


\section{Introduction}\label{section:intro}
  \IEEEPARstart{R}{eal} life processes are not immutable.
  Instead, they evolve to adapt to changes in their context, as new regulations or new consumption patterns.
  Changes can be planned by the organisation, but also happen unexpectedly.
  In the first case, the impact on the process can be computed and minimized.
  But in the second case, it may lead to wrong decisions because of outdated information.
  Thus, organizations should put in place prevention measures to detect when something is running differently from planned to reduce this negative impact.
  These unforeseen changes over time are known as concept drifts, which is one of the challenges presented in the \emph{Process Mining Manifesto}\cite{DBLP:conf/bpm/AalstAM11}.

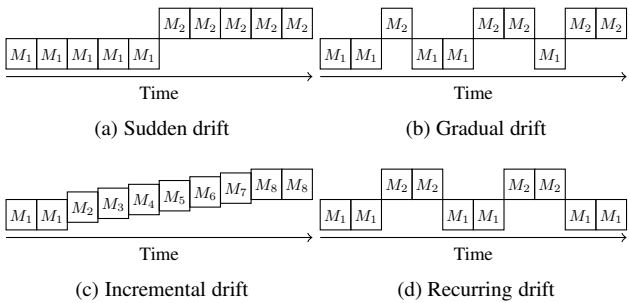
\begin{figure}[!t]
    \centering
    \subfloat[Sudden drift]{
      \resizebox{.5\columnwidth}{!}{

\begin{tikzpicture}
  \draw (0.0, 0.0) rectangle (0.4, 0.4);
  \draw (0.4, 0.0) rectangle (0.8, 0.4);
  \draw (0.8, 0.0) rectangle (1.2, 0.4);
  \draw (1.2, 0.0) rectangle (1.6, 0.4);
  \draw (1.6, 0.0) rectangle (2.0, 0.4);
  \draw (2.0, 0.4) rectangle (2.4, 0.8);
  \draw (2.4, 0.4) rectangle (2.8, 0.8);
  \draw (2.8, 0.4) rectangle (3.2, 0.8);
  \draw (3.2, 0.4) rectangle (3.6, 0.8);
  \draw (3.6, 0.4) rectangle (4.0, 0.8);

  \node at (0.2, 0.18) {\scalebox{.6}{$M_1$}};
  \node at (0.6, 0.18) {\scalebox{.6}{$M_1$}};
  \node at (1.0, 0.18) {\scalebox{.6}{$M_1$}};
  \node at (1.4, 0.18) {\scalebox{.6}{$M_1$}};
  \node at (1.8, 0.18) {\scalebox{.6}{$M_1$}};
  \node at (2.2, 0.58) {\scalebox{.6}{$M_2$}};
  \node at (2.6, 0.58) {\scalebox{.6}{$M_2$}};
  \node at (3.0, 0.58) {\scalebox{.6}{$M_2$}};
  \node at (3.4, 0.58) {\scalebox{.6}{$M_2$}};
  \node at (3.8, 0.58) {\scalebox{.6}{$M_2$}};

  \draw [->] (0,-0.1) -- node[below] {\scriptsize Time} ++ (4.0, 0.0);
\end{tikzpicture}
      }
      \label{figure:change-types:sudden}
    }
    \hspace*{-1.6em}
    \subfloat[Gradual drift]{
      \resizebox{.5\columnwidth}{!}{

\begin{tikzpicture}
  \draw (0.0, 0.0) rectangle (0.4, 0.4);
  \draw (0.4, 0.0) rectangle (0.8, 0.4);
  \draw (0.8, 0.4) rectangle (1.2, 0.8);
  \draw (1.2, 0.0) rectangle (1.6, 0.4);
  \draw (1.6, 0.0) rectangle (2.0, 0.4);
  \draw (2.0, 0.4) rectangle (2.4, 0.8);
  \draw (2.4, 0.4) rectangle (2.8, 0.8);
  \draw (2.8, 0.0) rectangle (3.2, 0.4);
  \draw (3.2, 0.4) rectangle (3.6, 0.8);
  \draw (3.6, 0.4) rectangle (4.0, 0.8);

  \node at (0.2, 0.18) {\scalebox{.6}{$M_1$}};
  \node at (0.6, 0.18) {\scalebox{.6}{$M_1$}};
  \node at (1.0, 0.58) {\scalebox{.6}{$M_2$}};
  \node at (1.4, 0.18) {\scalebox{.6}{$M_1$}};
  \node at (1.8, 0.18) {\scalebox{.6}{$M_1$}};
  \node at (2.2, 0.58) {\scalebox{.6}{$M_2$}};
  \node at (2.6, 0.58) {\scalebox{.6}{$M_2$}};
  \node at (3.0, 0.18) {\scalebox{.6}{$M_1$}};
  \node at (3.4, 0.58) {\scalebox{.6}{$M_2$}};
  \node at (3.8, 0.58) {\scalebox{.6}{$M_2$}};

  \draw [->] (0,-0.1) -- node[below] {\scriptsize Time} ++ (4.0, 0.0);
\end{tikzpicture}
      }
      \label{figure:change-types:gradual}
    }\\
    \subfloat[Incremental drift]{
      \resizebox{.5\columnwidth}{!}{

\begin{tikzpicture}
  \draw (0.0, 0.00) rectangle (0.4, 0.40);
  \draw (0.4, 0.00) rectangle (0.8, 0.40);
  \draw (0.8, 0.10) rectangle (1.2, 0.50);
  \draw (1.2, 0.15) rectangle (1.6, 0.55);
  \draw (1.6, 0.20) rectangle (2.0, 0.60);
  \draw (2.0, 0.25) rectangle (2.4, 0.65);
  \draw (2.4, 0.30) rectangle (2.8, 0.70);
  \draw (2.8, 0.35) rectangle (3.2, 0.75);
  \draw (3.2, 0.40) rectangle (3.6, 0.80);
  \draw (3.6, 0.40) rectangle (4.0, 0.80);

  \node at (0.2, 0.18) {\scalebox{0.6}{$M_1$}};
  \node at (0.6, 0.18) {\scalebox{0.6}{$M_1$}};
  \node at (1.0, 0.28) {\scalebox{0.6}{$M_2$}};
  \node at (1.4, 0.33) {\scalebox{0.6}{$M_3$}};
  \node at (1.8, 0.38) {\scalebox{0.6}{$M_4$}};
  \node at (2.2, 0.43) {\scalebox{0.6}{$M_5$}};
  \node at (2.6, 0.48) {\scalebox{0.6}{$M_6$}};
  \node at (3.0, 0.53) {\scalebox{0.6}{$M_7$}};
  \node at (3.4, 0.58) {\scalebox{0.6}{$M_8$}};
  \node at (3.8, 0.58) {\scalebox{0.6}{$M_8$}};

  \draw [->] (0,-0.1) -- node[below] {\scriptsize Time} ++ (4.0, 0.0);
\end{tikzpicture}
      }
      \label{figure:change-types:incremental}
    }
    \hspace*{-1.6em}
    \subfloat[Recurring drift]{
      \resizebox{.5\columnwidth}{!}{

\begin{tikzpicture}
  \draw (0.0, 0.0) rectangle (0.4, 0.4);
  \draw (0.4, 0.0) rectangle (0.8, 0.4);
  \draw (0.8, 0.4) rectangle (1.2, 0.8);
  \draw (1.2, 0.4) rectangle (1.6, 0.8);
  \draw (1.6, 0.0) rectangle (2.0, 0.4);
  \draw (2.0, 0.0) rectangle (2.4, 0.4);
  \draw (2.4, 0.4) rectangle (2.8, 0.8);
  \draw (2.8, 0.4) rectangle (3.2, 0.8);
  \draw (3.2, 0.0) rectangle (3.6, 0.4);
  \draw (3.6, 0.0) rectangle (4.0, 0.4);

  \node at (0.2, 0.18) {\scalebox{0.6}{$M_1$}};
  \node at (0.6, 0.18) {\scalebox{0.6}{$M_1$}};
  \node at (1.0, 0.58) {\scalebox{0.6}{$M_2$}};
  \node at (1.4, 0.58) {\scalebox{0.6}{$M_2$}};
  \node at (1.8, 0.18) {\scalebox{0.6}{$M_1$}};
  \node at (2.2, 0.18) {\scalebox{0.6}{$M_1$}};
  \node at (2.6, 0.58) {\scalebox{0.6}{$M_2$}};
  \node at (3.0, 0.58) {\scalebox{0.6}{$M_2$}};
  \node at (3.4, 0.18) {\scalebox{0.6}{$M_1$}};
  \node at (3.8, 0.18) {\scalebox{0.6}{$M_1$}};

  \draw [->] (0,-0.1) -- node[below] {\scriptsize Time} ++ (4.0, 0.0);
\end{tikzpicture}
      }
      \label{figure:change-types:recurring}
    }
    \caption{Types of concept drift based on their occurrence over time.}
    \label{figure:change-types}
  \end{figure}

  Changes can be classified based on their distribution over time \cite{DBLP:journals/csur/GamaZBPB14}:
  \begin{enumerate*}[label=(\roman*)]
    \item sudden drifts (\figurename~\ref{figure:change-types:sudden}), which means that the new concept replaces totally the previous one;
    \item gradual drifts (\figurename~\ref{figure:change-types:gradual}), where the new and the old concepts coexist for some time; and
    \item incremental drifts (\figurename~\ref{figure:change-types:incremental}), when the transition from the oldest concept to the newest one passes through some intermediate states that are, usually, some kind of combination from both.
  \end{enumerate*}
  Furthermore, when changes can be repeated over time, periodically switching between concepts, the change is classified as a recurrent drift (\figurename~\ref{figure:change-types:recurring}).
  In this paper, we focus on sudden drift detection.

  In addition, based on how data are processed\cite{DBLP:journals/tnn/BoseAZP14}, concept drift can be:
  \begin{enumerate*}[label=(\roman*)]
    \item offline, when change detection is made \emph{post-mortem}, being all data available from the beginning, and
    \item online, when change detection is made \emph{on-the-fly}, and new data are processed just when it is generated.
  \end{enumerate*}
  In this paper we focus on offline concept drift, which additionally faces two challenges:
  \begin{enumerate*}[label=\alph*)]
    \item the inherent complexity of process models, that can contain and combine different structures such as sequences, loops, parallel branches and choices; and
    \item the distinction between a change and an outlier, which is not always clear and may depend on the application domain.
  \end{enumerate*}

  Although, some authors have proposed different approaches for concept drift detection in process mining \cite{
    DBLP:conf/ida/CarmonaG12,
    DBLP:conf/iccsw/WeberBT11,
    DBLP:conf/www/JuniorTCCD18,
    DBLP:conf/otm/MaggiBCS13,
    DBLP:journals/tnn/BoseAZP14,
    DBLP:conf/bir/MartjushevBA15,
    DBLP:conf/apn/VTB15,
    DBLP:conf/IEEEscc/LiHWZC17,
    DBLP:conf/simpda/AccorsiS11,
    DBLP:conf/bpm/LuengoS11,
    DBLP:conf/simpda/HompesBADB15,
    DBLP:journals/tkde/MaaradjiDRO17,
    DBLP:conf/icde/LakshmananKD11,
    DBLP:conf/s-bpm-one/SeeligerNM17,
    DBLP:conf/otm/ZhengW017,
    DBLP:conf/er/OstovarMRHD16
  }, identifying all the possible change patterns\cite{DBLP:journals/dke/WeberRR08} with a short delay, allowing organizations to know exactly when the change took place and helping in the identification of the reasons that caused the change, is still a challenge.
  In addition, many approaches are unable to detect all change patterns, being this essential to reduce the cases when a change stays ignored to the organization.
  Another issue in some proposals is their high dependence on the end user, who is required to have some \emph{a-priori} knowledge of the process structure or skills to identify accurately the drift within a set of possibilities.

  In order to reduce the aforementioned issues, in this paper we present \proposedname, a novel and fully automatic approach based on discovery and conformance checking teckniques for offline detection of sudden concept drifts in the control-flow of process models.
  The method starts by defining a reference window, that will serve as a ground truth.
  Then traces are processed by a discovery algorithm to extract the corresponding process model.
  To process the remaining traces, the window is slided over the log, updating conformance metrics related to that process model.
  With these conformance values a regression is computed, and when the measurements decrease significantly a drift is detected.
  The underlying idea is that the value from the conformance metrics computed over the reference model and the new traces should decrease when the latter comes from a modified process, being this enough to determine if a change exists or not with a low delay.
  Specifically, the main contribution of this paper is the use of conformance metrics, in particular, fitness and precision, to detect changes in processes, which is a novel and unexplored approach so far.
  Namely, we propose the use of fitness metrics to detect changes that include traces with behavior not supported by the current process model, and the use of precision metrics to detect changes that implies behavior from the model disappearing from the real executions.

  \proposedname~has been tested using a dataset with 68 synthetic event logs.
  The results have been compared with the ones obtained by the methods available in the state of the art.
  \proposedname~has proved to be better at the accuracy level, getting better $F_{score}$ for all the event logs.
  In addition, \proposedname~gets very low delays, identifying changes closer to the point in which they really happened.
  Getting good values for both metrics is important for organizations for minimizing the number of unidentified changes and for reacting as soon as possible to those changes.

  The rest of this paper is structured as follows.
  In \sectionname~\ref{section:related-work} we analyze the main approaches to concept drift analysis.
  In \sectionname~\ref{section:preliminaries} we define a set of terms necessary to understand correctly our approach.
  In \sectionname~\ref{section:method} we detail our method for offline control-flow process concept drift detection.
  In \sectionname~\ref{section:validation} we present the experimentation performed to validate our approach and how it outperforms the main algorithms from the literature.
  Finally, in \sectionname~\ref{section:conclusions} we present our conclusions and outline our future work.

\section{Related Work}\label{section:related-work}
Although process mining is a rather active research field, concept drift analysis has not received much attention until recently.
It is worth noting that, although the method proposed in this paper focuses on offline detection, online approaches are also considered in the following analysis, because they can be easily adapted to detect this type of changes by simulating an online environment from the complete event log.

In \cite{DBLP:conf/ida/CarmonaG12}, authors propose a method for online concept drift detection using a polyhedron-based log representation.
Then, they monitor the probability that a trace falls into that polyhedron using the ADWIN algorithm \cite{DBLP:conf/sdm/BifetG07}.
The main drawback of this approach is that it can only detect the presence of a change, but it does not give any information about when it happened.

Online detection is also addressed in \cite{DBLP:conf/iccsw/WeberBT11}, where authors discover a probabilistic process model that, given an activity, assigns probability to every possible successors, and check how this probabilities evolve throughout the complete log using statistical hypothesis tests.
Although the method identifies drifts in most cases, small changes in less likely activities generate changes in the probabilities that can stay undetected.

In \cite{DBLP:conf/www/JuniorTCCD18} authors propose an online approach based on the extraction of histograms from traces, and then use a clustering algorithm to generate groups of similar traces.
A change is triggered when a new cluster appears.
An important drawback of this approach is that events order is not accounted for.
Thus, it can only detect the addition or removal of new activities, but not the changes in the precedence relations between them.

In \cite{DBLP:journals/tnn/BoseAZP14}, authors use a fixed-size window over some features extracted from the follows/precedes relations present in traces, and statistical hypothesis tests to evaluate whether these features have changed significantly.
The weak point of this method is that it requires a lot of interaction from the user, including previous knowledge of the process model and the areas where the changes can be located.
An extension of this work has been proposed in \cite{DBLP:conf/bir/MartjushevBA15}, where authors implement a recursive bisectioning approach.
Specifically, they take the traces that are involved in a drift detection and recursively split them in halves, with the aim of automatically localizing the change.
A drawback of this approach is that it still requires the user to know the possible changes in order to obtain good results.
A similar solution is presented in \cite{DBLP:conf/apn/VTB15}, where authors propose the usage of event class correlation as feature, and apply statistical hypothesis tests to detect changes.
However, it fails in detecting some change patterns such as the changes in the execution order of activities.

Another approach followed by some authors is the usage of clustering techniques in order to detect the drift.
In \cite{DBLP:conf/simpda/AccorsiS11}, authors cluster traces using the distance between pairs of activities.
However, this approach does not support models with loops.
Moreover, the distance can ignore certain change patterns depending on how many activities are affected by the change.
In \cite{DBLP:conf/bpm/LuengoS11}, authors extend a trace clustering algorithm\cite{DBLP:conf/bpm/BoseA09a} adding a time dimension in order to force clusters to include only consecutive traces, and thus be able to detect changes.
Their approach highly depends on the number of clusters, fixed by the user, and only obtain good results when the number of clusters is equal to the number of changes.
In \cite{DBLP:conf/simpda/HompesBADB15}, authors use a Markov clustering algorithm over different time windows in order to detect changes, but the approach does not focus
on the control-flow perspective.
Instead, multiple viewpoints of the process are taken into account simultaneously, mixing control-flow changes with behavioural and resources changes.

Another interesting approach, called \textit{ProcessDrift}, is proposed in \cite{DBLP:journals/tkde/MaaradjiDRO17}, where authors transform traces into \emph{partial-ordered-runs} and then apply a statistical hypothesis test over two windows (one for reference and one for detection) in order to detect changes.
The main drawback of this approach lies in its sensitivity to changes in the frequencies of certain relations present in the log, that may lead to false positives in the detection.
A related method is presented in \cite{DBLP:conf/er/OstovarMRHD16}, where authors focus on detecting the change at the event level instead of at trace level.
Specifically, they extract the $\alpha^+$ relations from two consecutive adaptive windows of events, and then, applying a statistical test, namely the G-test, compare the relations distribution of these two windows.
This allows the detection even with unfinished executions, and reduce the detection delay.
The drawback of this approach is that it requires high amounts of traces to be able to detect changes, being possible to ignore them when they are close from each other.

In \cite{DBLP:conf/icde/LakshmananKD11, DBLP:conf/s-bpm-one/SeeligerNM17} authors apply graph metrics to detect changes.
In \cite{DBLP:conf/icde/LakshmananKD11}, authors compare the eigenvectors and the eigenvalues of undirected weighted graphs representing the log at different instants.
In this graph, each vertex represents a trace.
The edges weight is the similarity between the vertex (traces) it connects.
However, this method needs a huge amount of traces, being unable to detect changes in logs with less than \num{2000} traces.
In \cite{DBLP:conf/s-bpm-one/SeeligerNM17} authors compare models over time using graph features, such as the node degree, the graph density or the occurrence of nodes and edges.
However, this approach does not perform well in processes with loops.

Finally, in \cite{DBLP:conf/otm/ZhengW017} authors present \textit{TPCDD}, a method that transforms the event log into a relation matrix using direct succession and weak order relations, where each column represents a trace and each row a relation.
Then, based on the trend of these relations generate candidate drift points.
These points are clustered using DBSCAN, in order to group candidates that belong to the same drift point.
This approach relies heavely in the user defining a correct radius for the DBSCAN algorithm, potentially getting a high number of false positives when it is too low and a high number of false negatives when it is too high.

With \proposedname~we take the aforementioned issues and try to minimize them in order to improve the results of the process drift detection.
The method removes any user interaction in the drift detection, requiring only a reference window size to be specified.
Moreover, the method is able to detect all change patterns independently of the process structure.
Furthermore, the method is designed to identify drifts with low delay, minimizing the detection of false negatives and positives.


\section{Preliminaries}\label{section:preliminaries}
  Below we present some concepts needed to understand the proposed method.
  The method takes an event log resulting of the executions of a process and tries to detect the changes in the execution of that process over time.

  \begin{definition}[\textbf{Event}]\label{definitions:event}
    An event $\varepsilon$ represents the execution of the activity $\alpha$ in the context of a process.
    Events have some mandatory attributes such as the activity, the execution case or the execution timestamp.
    They can also have optional attributes, such as the resource that performed the activity, the variables that were modified or the location.
  \end{definition}

  \begin{definition}[\textbf{Trace}]\label{definitions:trace}
    A trace is an ordered sequence of events $\tau = \langle \varepsilon_1, ..., \varepsilon_n \rangle$ where every event belongs to the same execution case.
  \end{definition}

  \begin{definition}[\textbf{Log}]\label{definitions:log}
    A log is defined as an ordered collection of traces $L = \langle \tau_1, ..., \tau_n \rangle$ where each trace represents one execution of the process.
    The size of the log, denoted as $|L|$, represents the number of traces in that log.
  \end{definition}

  A process model is a grah that describes the log behavior, that is, that allows to replay the log traces.
  A process model contains a representation of the coordination between the process activities, through sequences, parallels, choices and so on.
  In this paper we formalize process models using Petri nets.

  \begin{definition}[\textbf{Petri net}]\label{definitions:process-model}
    A Petri net is a tuple $N = (P, T, F)$, where:
    \begin{itemize}
      \item $P$ is a finite set of places
      \item $T$ is a finite set of transitions;
      \item $P \cap T = \emptyset$; and
      \item $F \subseteq (P \times T) \cup (T \times P)$ is a set of directed arcs.
    \end{itemize}
    Given $x \in T \cup P$, the set ${}^{\bullet} x = \{y\;|\;(y, x) \in F\}$ is the set of inputs of $x$, and $x^{\bullet} = \{y\;|\;(x, y) \in F\}$ the set of outputs of $x$.
    Given a Petri net $N = (P, T, F)$, a marking of $N$ is a mapping $M: P \rightarrow \mathbb{N}$, where $\mathbb{N}$ is the number of tokens in the place.

    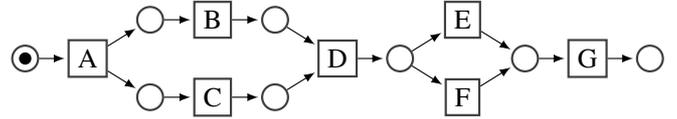
\begin{figure}[t]
      \centering
      \resizebox {\columnwidth} {!} {

\begin{tikzpicture}[node distance = 8mm, >=latex, auto]
  \tikzstyle{place} = [circle, thick, draw=black!75, minimum size=3mm]
  \tikzstyle{transition} = [rectangle, thick, draw=black!75, minimum size=4mm]

  \begin{scope}
    \node [place,tokens=1] (p1)  [] {};
    \node [place]          (p2)  [right of=p1, yshift=5mm,  xshift=8mm] {};
    \node [place]          (p3)  [right of=p1, yshift=-5mm, xshift=8mm] {};
    \node [place]          (p4)  [right of=p2, xshift=8mm] {};
    \node [place]          (p5)  [right of=p3, xshift=8mm] {};
    \node [place]          (p6)  [right of=p4, yshift=-5mm, xshift=8mm] {};
    \node [place]          (p7)  [right of=p6, xshift=8mm] {};
    \node [place]          (p8)  [right of=p7, xshift=8mm] {};

    \node [transition] (a) [right of=p1] {A}
      edge [pre]                            (p1)
      edge [post]                           (p2)
      edge [post]                           (p3);
    \node [transition] (b) [right of=p2] {B}
      edge [pre]                            (p2)
      edge [post]                           (p4);
    \node [transition] (c) [right of=p3] {C}
      edge [pre]                            (p3)
      edge [post]                           (p5);
    \node [transition] (d) [right of=p4, yshift=-5mm] {D}
      edge [pre]                            (p4)
      edge [pre]                            (p5)
      edge [post]                           (p6);
    \node [transition] (e) [right of=p6, yshift=5mm] {E}
      edge [pre]                            (p6)
      edge [post]                           (p7);
    \node [transition] (f) [right of=p6, yshift=-5mm] {F}
      edge [pre]                            (p6)
      edge [post]                           (p7);
    \node [transition] (g) [right of=p7] {G}
      edge [pre]                            (p7)
      edge [post]                           (p8);
  \end{scope}
\end{tikzpicture}
      }
      \caption{Example of a Petri net representing a process model.}
      \label{figure:example:petri-net}
    \end{figure}

    Processes usually have a unique start place $s \in P$ which has no inputs (${}^{\bullet} s = \emptyset$) and a unique end place $f \in P$ which has no outputs ($f^{\bullet} = \emptyset$).
    The initial marking of the Petri net $M_0$  contains only the initial place $M_{0}(s) = 1 \wedge \forall q \ne s \in P: M_{0}(q) = 0$.
    For a transition $t$ to be fired, all its input places must contain at least one token ($\forall p \in {}^{\bullet}t : M(p) \ge 1$).
    When $t$ is executed, it consumes a token from each of its inputs and puts a token in every of its outputs.
    Petri nets can be depicted as bipartite graphs, being transitions represented as rectangles and places as circles.
    A black bullet into a place represents a token.

    \figurename~\ref{figure:example:petri-net} shows a Petri net example.
    In this example, the process is conformed by the activities $A$, $B$, $C$, $D$, $E$, $F$ and $G$.
    In real executions, $A$ must be executed first.
    Then, $B$ and $C$ can be executed in any order.
    After this two activities are finished, $D$ is executed.
    Then, exclusively one of $E$ and $F$ must be executed.
    Finally, $G$ is executed and the process execution finishes.
  \end{definition}

  The quality of a process model $N$ with respect to a log $L$ can be estimated through some well established metrics such as fitness and precision.

  \begin{definition}[\textbf{Fitness metric}]\label{definitions:fitness}
    Given a log $L$ and a process model $N$, the fitness can be defined as a function $\gamma : L \times N \rightarrow \mathbb{R}$ which measures how well the log traces can be executed by the model.
  \end{definition}

  \begin{definition}[\textbf{Precision metric}]\label{definitions:precision}
    Given a log $L$ and a process model $N$, a precision measure can be defined as a function $\rho : L \times N \rightarrow \mathbb{R}$ which measures how much additional behaviour is allowed by the model that is never observed in the log.
  \end{definition}

  One of the objectives in this paper is the identification of changes in the process structure over time.
  This is done exploring the traces generated by the process and assessing if the more recent traces are product of the same process model.
  The concept of a sliding window captures this latter set of traces.

  \begin{definition}[\textbf{Sliding window}]\label{definitions:sliding-window}
    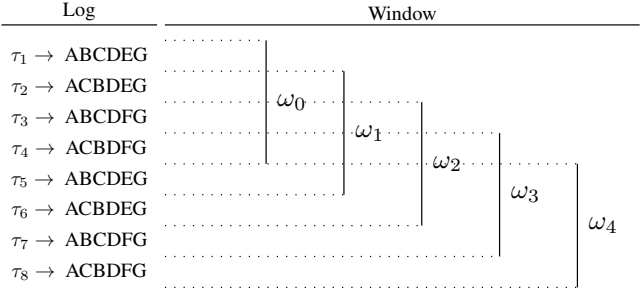
\begin{figure}[t]
      \centering
      \resizebox {\columnwidth} {!} {

\begin{tikzpicture}[scale=1.35]
  \draw [line width = .3pt] (-0.1, 1.7) -- node[above, inner sep=1pt] {\tiny Log} (.9, 1.7);
  \draw [line width = .3pt] (0.95, 1.7) -- node[above, inner sep=1pt] {\tiny Window} (4.0, 1.7);

  \node at (0.4, 1.5) {\scalebox{.5}{${\tau_{1} \rightarrow }$} \tiny ABCDEG};
  \node at (0.4, 1.3) {\scalebox{.5}{${\tau_{2} \rightarrow }$} \tiny ACBDEG};
  \node at (0.4, 1.1) {\scalebox{.5}{${\tau_{3} \rightarrow }$} \tiny ABCDFG};
  \node at (0.4, 0.9) {\scalebox{.5}{${\tau_{4} \rightarrow }$} \tiny ACBDFG};
  \node at (0.4, 0.7) {\scalebox{.5}{${\tau_{5} \rightarrow }$} \tiny ABCDEG};
  \node at (0.4, 0.5) {\scalebox{.5}{${\tau_{6} \rightarrow }$} \tiny ACBDEG};
  \node at (0.4, 0.3) {\scalebox{.5}{${\tau_{7} \rightarrow }$} \tiny ABCDFG};
  \node at (0.4, 0.1) {\scalebox{.5}{${\tau_{8} \rightarrow }$} \tiny ACBDFG};

  \draw [line width = .3pt] (1.6, 0.8) -- node[right, inner sep=1pt] {\resizebox{.035\hsize}{!}{$~\omega_0$}} (1.6, 1.6);
  \draw [line width = .3pt] (2.1, 0.6) -- node[right, inner sep=1pt] {\resizebox{.035\hsize}{!}{$~\omega_1$}} (2.1, 1.4);
  \draw [line width = .3pt] (2.6, 0.4) -- node[right, inner sep=1pt] {\resizebox{.035\hsize}{!}{$~\omega_2$}} (2.6, 1.2);
  \draw [line width = .3pt] (3.1, 0.2) -- node[right, inner sep=1pt] {\resizebox{.035\hsize}{!}{$~\omega_3$}} (3.1, 1.0);
  \draw [line width = .3pt] (3.6, 0.0) -- node[right, inner sep=1pt] {\resizebox{.035\hsize}{!}{$~\omega_4$}} (3.6, 0.8);

  \draw [dotted, line width= .3pt] (0.95, 1.6) -- (1.6, 1.6);
  \draw [dotted, line width= .3pt] (0.95, 1.4) -- (2.1, 1.4);
  \draw [dotted, line width= .3pt] (0.95, 1.2) -- (2.6, 1.2);
  \draw [dotted, line width= .3pt] (0.95, 1.0) -- (3.1, 1.0);
  \draw [dotted, line width= .3pt] (0.95, 0.8) -- (3.6, 0.8);
  \draw [dotted, line width= .3pt] (0.95, 0.6) -- (2.1, 0.6);
  \draw [dotted, line width= .3pt] (0.95, 0.4) -- (2.6, 0.4);
  \draw [dotted, line width= .3pt] (0.95, 0.2) -- (3.1, 0.2);
  \draw [dotted, line width= .3pt] (0.95, 0.0) -- (3.6, 0.0);
\end{tikzpicture}
      }
      \caption{Example of the sliding window behaviour.}
      \label{figure:example:window}
    \end{figure}

    Given a log $L$ and an integer $n \leq |L|$, a sliding window of size $n$ over the log $L$ can be defined as the sublog that at instant $i$ contains the last $n$ traces, denoted by $\omega_i = [\tau_i, ..., \tau_{i+n}]$.
    When a new trace is read from the log, $i$ is incremented, so the oldest trace from the window is forgotten, and the new one is added at the end of the window ($\omega_{i+1} = [\tau_{i+1}, ..., \tau_{i+n+1}]$).

    \figurename~\ref{figure:example:window} shows an example log and the behaviour of a four-sized sliding window over it.
    At instant $i=0$, the sliding window $\omega_0$ contains traces $\tau_1$ to $\tau_4$.
    When a new trace is read from the log, $i$ is incremented, so the oldest trace in the window ($\tau_1$) is forgotten and the new trace ($\tau_5$) is added to the window.
    This behaviour continues until the full log has been read.
  \end{definition}

  The structural evolution of the process over time, e.g., to adapt to new context circunstances or organization needs, is known as control-flow drift.

  \begin{definition}[\textbf{Control-flow drift}]\label{definitions:drift}
    Let $N_i = (P, T, F)$ and $N_j = (P^{*}, T^{*}, F^{*})$ be two process models at instants $i$ and $j = i + 1$, respectively.
    We say a control-flow drift exists when any of the following conditions is satisfied:
    \begin{itemize}
      \item $T \ne T^{*}$, which means that there are different activities in both models,
      \item $F \ne F^{*}$, which means that the relations between activities have changed.
    \end{itemize}
  \end{definition}

\section{\proposedname: Conformance Checking-based Concept Drift Detection}\label{section:method}
  The pseudocode of the proposed algorithm is listed in \algorithmname~\ref{algorithm:main}.
  Drift detection is based on a fixed-size sliding-window (definition \ref{definitions:sliding-window}), and thus the window size is the only parameter required to be specified by the user, aside from the event log.
  The rest of this paper assumes a window defined as a number of traces, but the method can be easily modified to allow windows based on a time interval.

  The algorithm starts by initializing a list $D$ of trace indices that produce a drift detection (line \ref{algorithm:lines:intialize-drifts}), and the initial window index $i = 0$ (line \ref{algorithm:lines:index-initialization}).
  Then, for this initial window, two lists of fitness and precision measurements $\Gamma$ and $\text{P}$ (lines \ref{algorithm:lines:fitness-initialization} and \ref{algorithm:lines:precision-initialization}), and two lists of drift candidates $D^{\Gamma}$ and $D^{\text{P}}$ (lines \ref{algorithm:lines:drift-fitness-initialization} and \ref{algorithm:lines:drift-precision-initialization}), one for each metric, are initialized.
  This two lists $D^{\Gamma}$ and $D^{\text{P}}$ contain a boolean for each processed window, indicating whether the window has been marked as a drift candidate or not.

  A reference model $N$ is discovered from the window content $\omega_i$ (line \ref{algorithm:lines:discovery}).
  \proposedname~is not tied to any discovery algorithm, although some features can affect its performance.
  First, the computational complexity since the discovery algorithm can be executed multiple times, one for each detected drift.
  Second, discovery algorithms that maximize fitness are more sensitive to small variations in traces, enhancing the subsequent detection of changes by traces replayability.
  This initialization phase is executed on the first trace window and when a new drift is detected by the algorithm.

  The detection step comprises lines \ref{algorithm:lines:detection-loop-start}-\ref{algorithm:lines:detection-loop-end}.
  As a first step, the fitness and precision metrics, $\gamma$ and $\rho$, are computed and appended to the respective list of measurements $\Gamma$ and $\text{P}$ (lines \ref{algorithm:lines:fitness-update} and \ref{algorithm:lines:precision-update}).
  Then, the window is classified as a drift candidate for each metric using the method described in \sectionname~\ref{section:method:drift-candidate-finding} and the result of this classification (either true or false) is stored in the corresponding list of drift candidates $D^{\Gamma}$ and $D^{\text{P}}$ (lines \ref{algorithm:lines:compute-fitness-drift} and \ref{algorithm:lines:compute-precision-drift}).
  Finally, the window index is incremented by one (i.e., the window slides one position), reading a new trace from the log (line \ref{algorithm:lines:slide-window}).
  A change is confirmed only if it persists in time.
  That means, the last $n$ windows has been classified as drift candidates either for fitness (lines \ref{algorithm:lines:confirmation-fitness-start}-\ref{algorithm:lines:confirmation-fitness-end}) or precision (lines \ref{algorithm:lines:confirmation-precision-start}-\ref{algorithm:lines:confirmation-precision-end}).
  This allows the method to prevent false positives due to the existence of temporal falls in the metrics caused by outlier traces.

  Once drift is confirmed, index $i$ is added to the list of confirmed drifts $D$ (line \ref{algorithm:lines:save-drift}), and the algorithm loops back to the initialization phase.
  Otherwise, detection step starts again with the new window, repeating until a change is detected or the log is completely analyzed.

  \begin{algorithm}[t]
    \caption{Conformance Checking-based Drift Detection}
    \label{algorithm:main}

\begin{algorithmic}[1]
  \Statex \textbf{Inputs:} an event log  $L$ and a window size $n < |L|$
  \Statex \textbf{Outputs:} a list of trace indices causing drift $\text{D}$
  \Statex
  \Procedure{ConceptDriftDetection}{$L, n$}
    \State $\text{D} \gets [~]$
    \label{algorithm:lines:intialize-drifts}
    \State $\mathrlap{i}\hphantom{\text{D}} \gets 0$
    \label{algorithm:lines:index-initialization}

    \While{$i \leq |L| - n$}
      \State $\mathrlap{\Gamma}\hphantom{\text{D}^{\Gamma}} \gets [~]$
      \Comment{fitness measures (\definitionname~\ref{definitions:fitness})}
      \label{algorithm:lines:fitness-initialization}
      \State $\mathrlap{\text{P}}\hphantom{\text{D}^{\Gamma}} \gets [~]$
      \Comment{precision measures (\definitionname~\ref{definitions:precision})}
      \label{algorithm:lines:precision-initialization}
      \State $\text{D}^{\Gamma} \gets [~]$
      \Comment{drift candidates (fitness)}
      \label{algorithm:lines:drift-fitness-initialization}
      \State $\mathrlap{\text{D}^{\text{P}}}\hphantom{\text{D}^{\Gamma}} \gets [~]$
      \Comment{drift candidates (precision)}
      \label{algorithm:lines:drift-precision-initialization}
      \State $\mathrlap{\omega_i}\hphantom{\text{D}^{\Gamma}} \gets \{L_{i}, ..., L_{i+n}\}$
      \State $\mathrlap{\text{N}}\hphantom{\text{D}^{\Gamma}} \gets discover(\omega_i)$
      \label{algorithm:lines:discovery}

      \While{$(i \leq (|L| - n)) \wedge \hphantom{meteespacioaquiquenecesito} \linebreak \hphantom{procwhile-space} \neg $ \Call{ConfirmDrift}{$n, \text{D}^{\Gamma}, \text{D}^{\text{P}}$}}
      \label{algorithm:lines:detection-loop-start}
        \State $\mathrlap{\Gamma}\hphantom{\text{D}^{\Gamma}} \gets \Gamma :: \gamma(\omega_i, M)$
        \Comment{append current fitness}
        \label{algorithm:lines:fitness-update}
        \State $\mathrlap{\text{P}}\hphantom{\text{D}^{\Gamma}} \gets \mathrlap{\text{P}}\hphantom{\Gamma} ::  \rho(\omega_i, M)$
        \Comment{append current precision}
        \label{algorithm:lines:precision-update}
        \State $\text{D}^{\Gamma} \gets \text{D}^{\Gamma} :: $ \Call{IdentifyDrift}{$n, \Gamma, \text{D}^{\Gamma}$}
        \label{algorithm:lines:compute-fitness-drift}
        \State $\mathrlap{\text{D}^{\text{P}}}\hphantom{\text{D}^{\Gamma}} \gets \mathrlap{\text{D}^{\text{P}}}\hphantom{\text{D}^{\Gamma}} :: $ \Call{IdentifyDrift}{$n, \text{P}, \text{D}^{\text{P}}$}
        \label{algorithm:lines:compute-precision-drift}
        \State $\mathrlap{i}\hphantom{\text{D}^{\Gamma}} \gets i+1$
        \label{algorithm:lines:slide-window}
      \EndWhile
      \label{algorithm:lines:detection-loop-end}

      \If{\Call{ConfirmDrift}{$n, \text{D}^{\Gamma}, \text{D}^{\text{P}}$}}
      \label{algorithm:lines:confirm-drift}
        \State $\text{D} \gets \text{D} ::  i$
        \label{algorithm:lines:save-drift}
      \EndIf
    \EndWhile

    \State \textbf{return} $\text{D}$
  \EndProcedure

  \Statex

  \Function{IdentifyDrift}{$n, data, \text{D}^{*}$}
  \label{algorithm:lines:detection:start}
    \State $\mathrlap{\Upsilon}\hphantom{m^{<}} \gets$ \Call{$\Theta$}{[$data_{|data| - (n/2)}, ..., data_{|data|}]$}
    \label{algorithm:lines:detection:regression}
    \State $m^{<} \gets m(\Upsilon) < 0 \wedge p(\Upsilon) < 0.05$
    \label{algorithm:lines:detection:slope:negative}
    \State $m^{>} \gets m(\Upsilon) > 0 \wedge p(\Upsilon) < 0.05$
    \label{algorithm:lines:detection:slope:positive}
    \State $m^{=} \gets (\neg\,m^{<}) \wedge (\neg\,m^{>}) $
    \label{algorithm:lines:detection:slope:zero}
    \State \textbf{return} $(|data| > n/2) \wedge (m^{<} \vee m^{>} \vee (m^{=} \wedge (\text{D}^{*}_{|\text{D}^{*}|} = true))) $
  \EndFunction
  \label{algorithm:lines:detection:end}

  \Statex

  \Function{ConfirmDrift}{$n, \text{D}^{\Gamma}, \text{D}^{\text{P}}$}
  \label{algorithm:lines:confirmation-start}
    \If{$\{\forall i \in [|\text{D}^{\Gamma}| - n, |\text{D}^{\Gamma}|] : \text{D}^{\Gamma}_i = true\}$}
    \label{algorithm:lines:confirmation-fitness-start}
      \State $d^{\Gamma} \gets true$
    \Else
      \State $d^{\Gamma} \gets false$
    \EndIf
    \label{algorithm:lines:confirmation-fitness-end}
    \If{$\{\forall i \in [|\text{D}^{\text{P}}| - n, |\text{D}^{\text{P}}|] : \text{D}^{\text{P}}_i = true\}$}
    \label{algorithm:lines:confirmation-precision-start}
      \State $d^{\text{P}} \gets true$
    \Else
      \State $d^{\text{P}} \gets false$
    \EndIf
    \label{algorithm:lines:confirmation-precision-end}

    \State \textbf{return} $(|\text{D}^{\Gamma}| \geq n \wedge d^{\Gamma}) \vee (|\text{D}^{\text{P}}| \geq n \wedge d^{\text{P}})$
  \EndFunction
  \label{algorithm:lines:confirmation-end}
\end{algorithmic}

  \end{algorithm}

  \subsection{Drift Detection}\label{section:method:drift-candidate-finding}
    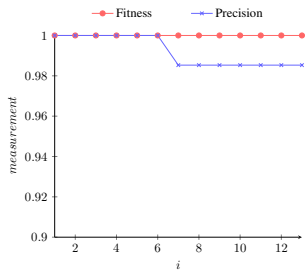
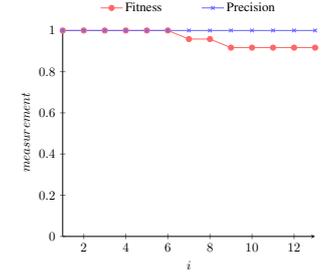
\begin{figure}[t]
      \centering
      \subfloat[Model $N_1$. ]{
        \resizebox {.97\linewidth} {!} {

\begin{tikzpicture}[node distance = 8mm, >=latex, auto]
  \tikzstyle{place} = [circle, thick, draw=black!75, minimum size=3mm]
  \tikzstyle{transition} = [rectangle, thick, draw=black!75, minimum size=4mm]

  \begin{scope}
    \node [place,tokens=1] (p1)  [] {};
    \node [place]          (p2)  [right of=p1, yshift=5mm,  xshift=8mm] {};
    \node [place]          (p3)  [right of=p1, yshift=-5mm, xshift=8mm] {};
    \node [place]          (p4)  [right of=p2, xshift=8mm] {};
    \node [place]          (p5)  [right of=p3, xshift=8mm] {};
    \node [place]          (p6)  [right of=p4, yshift=-5mm, xshift=8mm] {};
    \node [place]          (p7)  [right of=p6, xshift=8mm] {};
    \node [place]          (p8)  [right of=p7, xshift=8mm] {};

    \node [transition] (a) [right of=p1] {A}
      edge [pre]                            (p1)
      edge [post]                           (p2)
      edge [post]                           (p3);
    \node [transition] (b) [right of=p2] {B}
      edge [pre]                            (p2)
      edge [post]                           (p4);
    \node [transition] (c) [right of=p3] {C}
      edge [pre]                            (p3)
      edge [post]                           (p5);
    \node [transition] (d) [right of=p4, yshift=-5mm] {D}
      edge [pre]                            (p4)
      edge [pre]                            (p5)
      edge [post]                           (p6);
    \node [transition] (e) [right of=p6, yshift=5mm] {E}
      edge [pre]                            (p6)
      edge [post]                           (p7);
    \node [transition] (f) [right of=p6, yshift=-5mm] {F}
      edge [pre]                            (p6)
      edge [post]                           (p7);
    \node [transition] (g) [right of=p7] {G}
      edge [pre]                            (p7)
      edge [post]                           (p8);
  \end{scope}
\end{tikzpicture}
        }
        \label{figure:change-detection:model1}
      }
      \\
      \subfloat[Model $N_2$.]{
        \resizebox {.97\linewidth} {!} {

\begin{tikzpicture}[node distance = 8mm, >=latex, bend angle=45, auto]
  \tikzstyle{place} = [circle, thick, draw=black!75, minimum size=3mm]
  \tikzstyle{transition} = [rectangle, thick, draw=black!75, minimum size=4mm]
  \tikzstyle{silent} = [rectangle, thin, draw=black!75, fill=black!75, minimum width=1mm, minimum height=4mm, inner sep=0pt]

  \node [place,tokens=1] (p1)  [] {};
  \node [place]          (p2)  [right of=p1, xshift=8mm] {};
  \node [place]          (p3)  [right of=p2, xshift=8mm] {};
  \node [place]          (p4)  [right of=p3, xshift=8mm] {};
  \node [place]          (p5)  [right of=p4, xshift=8mm] {};
  \node [place]          (p6)  [right of=p5, xshift=8mm] {};
  \node [place]          (p7)  [right of=p6, xshift=8mm] {};

  \node [transition] (a) [right of=p1] {A}
    edge [pre]                            (p1)
    edge [post]                           (p2);
  \node [transition] (b) [right of=p2] {B}
    edge [pre]                            (p2)
    edge [post]                           (p3);
  \node [transition] (c) [right of=p3] {C}
    edge [pre]                            (p3)
    edge [post]                           (p4);
  \node [transition] (d) [right of=p4] {D}
    edge [pre]                            (p4)
    edge [post]                           (p5);
  \node [transition] (e) [right of=p5, yshift=5mm] {E}
    edge [pre]                            (p5)
    edge [post]                           (p6);
  \node [transition] (f) [right of=p5, yshift=-5mm] {F}
    edge [pre]                            (p5)
    edge [post]                           (p6);
  \node [transition] (g) [right of=p6] {G}
    edge [pre]                            (p6)
    edge [post]                           (p7);
\end{tikzpicture}
        }
        \label{figure:change-detection:model2}
      }
      \\
      \subfloat[Log generated using $N_1$ as the initial process and $N_2$ as the modified one.]{
        \resizebox {.45\columnwidth} {!} {
\tabulinesep=2pt
\begin{tabu}{cp{1cm}l}
  ID           & \multicolumn{2}{c}{Trace} \\ \tabucline[.8pt black]{-}
  $\tau_{1}$   & & A B C D E G \\
  $\tau_{2}$   & & A C B D E G \\
  $\tau_{3}$   & & A B C D F G \\
  $\tau_{4}$   & & A C B D F G \\
  $\tau_{5}$   & & A B C D E G \\
  $\tau_{6}$   & & A C B D E G \\
  $\tau_{7}$   & & A B C D F G \\
  $\tau_{8}$   & & A C B D F G \\ \tabucline[.4pt on 1pt black!60 off 4pt]{-}
  $\tau_{9}$   & & A B C D E G \\
  $\tau_{10}$  & & A B C D F G \\
  $\tau_{11}$  & & A B C D E G \\
  $\tau_{12}$  & & A B C D F G \\
  $\tau_{13}$  & & A B C D E G \\
  $\tau_{14}$  & & A B C D F G \\
  $\tau_{15}$  & & A B C D E G \\
  $\tau_{16}$  & & A B C D F G \\
\end{tabu}

        }
        \label{figure:change-detection:log1}
      }
      \hfill
      \subfloat[Log generated using $N_2$ as the initial process and $N_1$ as the modified one.]{
        \resizebox {.45\columnwidth} {!} {
\tabulinesep=2pt
\begin{tabu}{cp{1cm}l}
  ID           & \multicolumn{2}{c}{Trace} \\ \tabucline[.8pt black]{-}
  $\tau_{1}$   & & A B C D E G \\
  $\tau_{2}$   & & A B C D F G \\
  $\tau_{3}$   & & A B C D E G \\
  $\tau_{4}$   & & A B C D F G \\
  $\tau_{5}$   & & A B C D E G \\
  $\tau_{6}$   & & A B C D F G \\
  $\tau_{7}$   & & A B C D E G \\
  $\tau_{8}$   & & A B C D F G \\ \tabucline[.4pt on 1pt black!60 off 4pt]{-}
  $\tau_{9}$   & & A B C D E G \\
  $\tau_{10}$  & & A C B D E G \\
  $\tau_{11}$  & & A B C D F G \\
  $\tau_{12}$  & & A C B D F G \\
  $\tau_{13}$  & & A B C D E G \\
  $\tau_{14}$  & & A C B D E G \\
  $\tau_{15}$  & & A B C D F G \\
  $\tau_{16}$  & & A C B D F G
\end{tabu}

        }
        \label{figure:change-detection:log2}
      }
      \\
      \subfloat[Fitness and precision evolution when using a window of size 4, log from \figurename~\ref{figure:change-detection:log1} and reference model $N_1$]{
        \resizebox {.46\columnwidth} {!} {

\begin{tikzpicture}
	\begin{axis}[
		xlabel=$i$,
		ylabel={$measurement$},
		xmin = 1,
		xmax = 13,
		ymin = 0.9,
		ymax = 1.001,
		legend cell align={right},
		legend style={at={(0.5, 1.15)}, anchor=north, draw=none, /tikz/every even column/.append style={column sep=10mm}},
		legend columns=-1,
		axis y line=left,
    axis y line*=left,
		axis x line*=bottom,
    axis x line=bottom,
	]
	  \addplot[color=red!60,mark=*] coordinates {
	    (1,  1.0)
	    (2,  1.0)
	    (3,  1.0)
	    (4,  1.0)
	    (5,  1.0)
	    (6,  1.0)
	    (7,  1.0)
	    (8,  1.0)
	    (9,  1.0)
	    (10, 1.0)
	    (11, 1.0)
	    (12, 1.0)
	    (13, 1.0)
		};

	  \addplot[color=blue!60,mark=x] coordinates {
	    (1,  1.0)
	    (2,  1.0)
	    (3,  1.0)
	    (4,  1.0)
	    (5,  1.0)
	    (6,  1.0)
	    (7,  0.9852941145856172)
	    (8,  0.9852941145856172)
	    (9,  0.9852941193389075)
	    (10, 0.9852941193389075)
	    (11, 0.9852941193389075)
	    (12, 0.9852941193389075)
	    (13, 0.9852941193389075)
		};
		\legend{Fitness, Precision}
	\end{axis}
\end{tikzpicture}
        }
        \label{figure:change-detection:log1:measurements}
      }
      \hfill
      \subfloat[Fitness and precision evolution when using a window of size 4, log from \figurename~\ref{figure:change-detection:log2} and reference model $N_2$.]{
        \resizebox {.46\columnwidth} {!} {

\begin{tikzpicture}
	\begin{axis}[
		xlabel=$i$,
		ylabel={$measurement$},
		xmin = 1,
		xmax = 13,
		ymin = 0.0,
		ymax= 1.01,
		legend style={at={(0.5, 1.15)}, anchor=north, draw=none, /tikz/every even column/.append style={column sep=10mm}},
		legend columns=-1,
		axis y line=left,
    axis y line*=left,
		axis x line*=bottom,
    axis x line=bottom,
	]
	  \addplot[color=red!60,mark=*] coordinates {
	    (1,  1.0)
	    (2,  1.0)
	    (3,  1.0)
	    (4,  1.0)
	    (5,  1.0)
	    (6,  1.0)
	    (7,  0.9583333333333334)
	    (8,  0.9583333333333334)
	    (9,  0.9166666666666667)
	    (10, 0.9166666666666667)
	    (11, 0.9166666666666667)
	    (12, 0.9166666666666667)
	    (13, 0.9166666666666667)
		};

	  \addplot[color=blue!60,mark=x] coordinates {
	    (1,  1.0)
	    (2,  1.0)
	    (3,  1.0)
	    (4,  1.0)
	    (5,  1.0)
	    (6,  1.0)
	    (7,  1.0)
	    (8,  1.0)
	    (9,  1.0)
	    (10, 1.0)
	    (11, 1.0)
	    (12, 1.0)
	    (13, 1.0)
		};
		\legend{Fitness, Precision}
	\end{axis}
\end{tikzpicture}
        }
        \label{figure:change-detection:log2:measurements}
      }
      \caption{Measurements evolution for two logs with different changes.}
      \label{figure:change-detection}
    \end{figure}

    Fitness and precision separately can not detect all possible changes, but a combination of both can.
    This is illustrated in the example depicted in \figurename~\ref{figure:change-detection}.
    \figurename~\ref{figure:change-detection:model1} and \figurename~\ref{figure:change-detection:model2} present two models, henceforth $N_1$ and $N_2$.
    The difference between both models is that activities $B$ and $C$ that are in parallel in $N_1$, while they are in sequence in $N_2$.
    Let suppose that a process changes from the model $N_1$ to $N_2$ at instant $i=8$, which log is represented in \figurename~\ref{figure:change-detection:log1}, henceforth denoted as $L_1$.
    Traces $\tau_{1}$ to $\tau_{8}$ correspond to the execution of $N_1$ and traces $\tau_{9}$ to $\tau_{16}$ correspond to $N_2$.
    In $L_1$, the concurrent execution of activities $B$ and $C$ becomes a sequence from $\tau_{9}$ onwards.
    With this change there is not any trace that is not replayable by the model, so the fitness remains unaltered.
    However, no trace in the window contains the path $A\rightarrow C\rightarrow D$ from $\tau_{9}$ onwards, so the precision falls.
    This can be seen in \figurename~\ref{figure:change-detection:log1:measurements}, where precision falls because the model allows more behaviour than is present in the traces, but fitness remains unaltered.

    Let also suppose a different change, from model $N_2$ to $N_1$, at the same time instant, which log is represented in \figurename~\ref{figure:change-detection:log2}, henceforth denoted as $L_2$.
    Traces $\tau_{1}$ to $\tau_{8}$ are generated by $N_2$ and traces $\tau_{9}$ to $\tau_{16}$ by $N_1$.
    In $L_2$, $B$ and $C$, originally in sequence, are in parallel from $\tau_{9}$ onwards.
    This change can not be detected using precision (the model does not generate more behavior than the present in the log), but it can be detected though fitness, since $\tau_{10}$, $\tau_{12}$, $\tau_{14}$ and $\tau_{16}$ can not be replayed by $N_2$.
    This situation is represented in \figurename~\ref{figure:change-detection:log2:measurements}, where precision remains unchanged, but fitness falls in the \ordinalnum{7} iteration of the algorithm.
    This detection mechanism is listed in lines \ref{algorithm:lines:detection:start} to \ref{algorithm:lines:detection:end} of \algorithmname~\ref{algorithm:main} (function \textsc{ConfirmDrift}).
    As a first step, a linear regression is computed over the last $n/2$ measurements for both fitness and precision (line \ref{algorithm:lines:detection:regression}).
    A statistical test is performed to ensure that results are statistically significant, determining the probability of the regression slope being \num{0} with a $p_{value}$ of \num{0.05}.
    When $H_0$ is rejected (i.e. $p(\Upsilon) < 0.05$) we asume that enough evidences exist to accept the slope value $m(\Upsilon)$.
    Otherwise, we can not asume that the slope value is different from \num{0}.
    There are three possible situations:
    \begin{enumerate}
      \item The regression slope is negative (line \ref{algorithm:lines:detection:slope:negative}): metrics get lower values, so more traces are not replayable for fitness or, conversely, more paths of the model are not contained in traces for precision.
      Thus the window is marked as a drift candidate.
      \item The regression slope is positive (line \ref{algorithm:lines:detection:slope:positive}): metrics get higher values, since more traces are replayable for fitness or, conversely, more paths of the model are contained in traces for precision.
      Thus the window is marked as a drift candidate.
      \item The regression slope is zero (line \ref{algorithm:lines:detection:slope:zero}): no change in conformance metrics, i.e., the window does not present any drift. In this case, a drift can also be detected, but only if the previous window was marked as a drift candidate.
    \end{enumerate}

    \begin{figure}[t]
      \centering
      \subfloat[Log 1 (\texttt{pl}). Precision falls every time a change happens but fitness remain unaltered.]{
        \resizebox {.97\linewidth} {!} {

\begin{tikzpicture}
  \begin{scope}
    \begin{axis}[
      width=9cm,
      height=4cm,
      axis y line=left,
      axis y line*=left,
      axis x line*=bottom,
      axis x line=bottom,
      ymin = 0.8,
      ymax = 1.05,
      ytick = \empty,
      xtick = \empty,
      ylabel = {Fitness}
    ]
      \addplot [blue, thick] table [x=index, y=value, col sep=comma, mark=none] {figures/fitness-pl.csv};
    \end{axis}
  \end{scope}

  \begin{scope} [xshift=8cm]
    \begin{axis}[
      width=9cm,
      height=4cm,
      axis y line=left,
      axis y line*=left,
      axis x line*=bottom,
      axis x line=bottom,
      ymin = 0.8,
      ymax = 1.05,
      ytick = \empty,
      xtick = \empty,
      ylabel = {Precision}
    ]
      \addplot [red, thick] table [x=index, y=value, col sep=comma, mark=none] {figures/precision-pl.csv};
    \end{axis}
  \end{scope}
\end{tikzpicture}
        }
        \label{figure:validation:metrics-example:pl}
      }\\
      \subfloat[Log 2 (\texttt{cb}). Fitness falls but precision remains unaltered in every change.]{
        \resizebox {0.97\linewidth} {!} {

\begin{tikzpicture}
  \begin{scope}
    \begin{axis}[
      width=9cm,
      height=4cm,
      axis y line=left,
      axis y line*=left,
      axis x line*=bottom,
      axis x line=bottom,
      ymin = 0.8,
      ymax = 1.05,
      ytick = \empty,
      xtick = \empty,
      ylabel = {Fitness}
    ]
      \addplot [blue, thick] table [x=index, y=value, col sep=comma, mark=none] {figures/fitness-cb.csv};
    \end{axis}
  \end{scope}

  \begin{scope} [xshift=8cm]
    \begin{axis}[
      width=9cm,
      height=4cm,
      axis y line=left,
      axis y line*=left,
      axis x line*=bottom,
      axis x line=bottom,
      ymin = 0.8,
      ymax = 1.05,
      ytick = \empty,
      xtick = \empty,
      ylabel = {Precision}
    ]
      \addplot [red, thick] table [x=index, y=value, col sep=comma, mark=none] {figures/precision-cb.csv};
    \end{axis}
  \end{scope}
\end{tikzpicture}
        }
        \label{figure:validation:metrics-example:cb}
      }
      \caption{Evolution of fitness and precision metrics when computed with the last 100 traces on logs \texttt{pl} and \texttt{cb}.}
      \label{figure:validation:metrics-example}
    \end{figure}
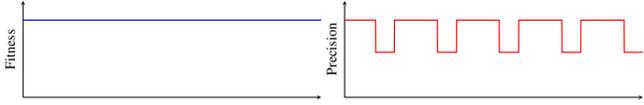
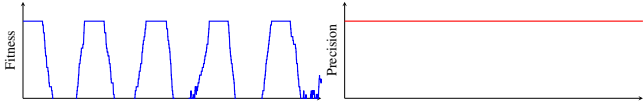

    We can see the negative regression slopes in precision and fitness, for the aforementioned logs, in \figurename~\ref{figure:change-detection:log1:measurements} and \figurename~\ref{figure:change-detection:log2:measurements}, respectively.
    A more complete example with several changes is depicted in \figurename~\ref{figure:validation:metrics-example}.
    Specifically, this example shows the drift detection using  the logs \texttt{pl} and \texttt{cb}, that will be described in \sectionname~\ref{section:validation:data}, which contain a change every 250 traces.
    In the case of \texttt{pl} (\figurename~\ref{figure:validation:metrics-example:pl}), two fragments that are originaly executed in a concurrent form are transformed into a sequential execution, which should imply a reduction in precision but not in fitness.
    In the case of \texttt{cb} (\figurename~\ref{figure:validation:metrics-example:cb}), a fragment is transformed from mandatory to skipable, which should imply a reduction in fitness but not in precision.
    When the regression is computed with few data the slope can oscillate, producing false positives.
    To prevent this effect, the algorithm only checks the previous conditions if there are at least $n/2$ measurements available.

  \subsection{Custom Fitness and Precision}\label{section:validation:metrics}
    Traditional fitness and precision metrics are designed to assess the global quality of a model.
    However, \proposedname~use them to detect structural changes in the execution of a process.
    The main intuitions behind this hypothesis are:
    \begin{enumerate}
      \item Fitness can measure how well the log can be replayed by the model.
      When new activities are added, removed or the precedence is modified, this metric will fall, indicating a change.
      \item Precision can measure how well the model represents the behavior that is present in the log.
      When a path no longer exists, the precision values will fall, indicating a change.
    \end{enumerate}
    Moreover, we also propose two simpler fitness and precision metrics aside from the well-established metrics from the state of the art \cite{DBLP:journals/is/RozinatA08, DBLP:journals/tkde/BrouckeWVB14}.
    These two approaches have much lower computational complexity and were designed to detect changes in simple and noise-free logs.
    In the case of fitness, we use the percentage of replayable traces.
    This approach is not particularly useful for measuring the quality of a model, since it equally penalizes traces that do not fit the model and those that deviate slightly from it.
    Despite this, it can be used to estimate changes in fitness, since a change in the percentage of traces that can be replayed in the model always leads to a change in the metric value.
    For precision, the following approach is used:
    \begin{equation}
      PC = 1 - \frac{ \vert OLP \setminus DFR \vert}{ \vert OLP \vert }
      \label{equation:model-precision}
    \end{equation}
    where:
    \begin{itemize}
      \item A set of one-length paths ($OLP$) is extracted from the model.
      An $OLP$ is a pair of activities that are directly connected in the process model, without any other activity in between
      \item A set of directly-follows relations ($DFR$) is extracted from the log.
      A $DFR$ is a pair os activities that appear one after the other in the log, without any activity in between.
    \end{itemize}

    Eq. (\ref{equation:model-precision}) does not measure the precision \textit{per se}, but the change in the precision.
    The moment a $OLP$ stops appearing in the log is indicative that some path of the model has disappeared.
    The proposed approach returns \num{1} when all the supported behavior of the model appear at least once in the log, and \num{0} otherwise, i.e., when none of the behavior supported by the model appears in the log.
    The computation of this metric is illustrated with an example in \figurename~\ref{figure:validation:metrics:precision-change}.

    \begin{figure}[t]
      \centering
      \subfloat{
        \resizebox {.97\linewidth} {!} {

\begin{tikzpicture}
  \tikzstyle{place} = [circle, thick, draw=black!75, minimum size=3mm]
  \tikzstyle{transition} = [rectangle, thick, draw=black!75, minimum size=4mm]

  \begin{scope}[node distance = 8mm, >=latex, auto]
    \node [place,tokens=1] (p1)  [] {};
    \node [place]          (p2)  [right of=p1, yshift=5mm,  xshift=8mm] {};
    \node [place]          (p3)  [right of=p1, yshift=-5mm, xshift=8mm] {};
    \node [place]          (p4)  [right of=p2, xshift=8mm] {};
    \node [place]          (p5)  [right of=p3, xshift=8mm] {};
    \node [place]          (p6)  [right of=p4, yshift=-5mm, xshift=8mm] {};
    \node [place]          (p7)  [right of=p6, xshift=8mm] {};
    \node [place]          (p8)  [right of=p7, xshift=8mm] {};

    \node [transition] (a) [right of=p1] {A}
      edge [pre]                            (p1)
      edge [post]                           (p2)
      edge [post]                           (p3);
    \node [transition] (b) [right of=p2] {B}
      edge [pre]                            (p2)
      edge [post]                           (p4);
    \node [transition] (c) [right of=p3] {C}
      edge [pre]                            (p3)
      edge [post]                           (p5);
    \node [transition] (d) [right of=p4, yshift=-5mm] {D}
      edge [pre]                            (p4)
      edge [pre]                            (p5)
      edge [post]                           (p6);
    \node [transition] (e) [right of=p6, yshift=5mm] {E}
      edge [pre]                            (p6)
      edge [post]                           (p7);
    \node [transition] (f) [right of=p6, yshift=-5mm] {F}
      edge [pre]                            (p6)
      edge [post]                           (p7);
    \node [transition] (g) [right of=p7] {G}
      edge [pre]                            (p7)
      edge [post]                           (p8);
  \end{scope}
\end{tikzpicture}
        }
      }\\
      \subfloat{
        \resizebox {.25\linewidth} {!} {
\tabulinesep=2pt
\begin{tabu}{cXc}
  \textbf{Log} \\ \toprule
  A B C D E G  \\
  A C B D E G  \\
  A B C D E G  \\
  A C B D E G  \\
  A C B D E G  \\
  A B C D E G  \\
  A B C D E G  \\
  A C B D E G  \\
  A B C D E G  \\ \bottomrule
\end{tabu}

        }
      }\hfill
      \subfloat{
        \resizebox {.65\linewidth} {!} {
          \input{figures/precision-change-example-compute}
        }
      }
      \caption{$PC$ computation example.}
      \label{figure:validation:metrics:precision-change}
    \end{figure}

\section{Experimentation}\label{section:validation}
  Concept drift algorithms are assessed based on two quality measures: $F_{score}$ (\ref{equation:fscore}), which is an accuracy metric computed as the harmonic mean between precision (\ref{equation:precision}) and recall (\ref{equation:recall}); and delay (henceforth $\Delta$), which is the distance between the point when the change really happened and when it is detected.

  \begin{subequations}
    \begin{equation}
      F_{score} = \frac{2 \times precision \times recall}{precision + recall}
      \label{equation:fscore}
    \end{equation}
    \begin{equation}
      precision = \frac{TP}{TP + FP}
      \label{equation:precision}
    \end{equation}
    \begin{equation}
      recall = \frac{TP}{TP + FN}
      \label{equation:recall}
    \end{equation}
  \end{subequations}

\begin{figure}[t]
    \centering
    \resizebox {\linewidth} {!} {

\begin{tikzpicture}
  \filldraw[pattern = north west lines, pattern color=black!40!white, draw=black!40!white] (0, 0.20)  rectangle (10,-0.20);
  \filldraw[pattern = north west lines, pattern color=black!40!white, draw=black!40!white] (15, 0.20) rectangle (25,-0.20);

  \draw (0,0) -- (25, 0);

  \draw (0,0.25)  -- (0,-0.25);
  \draw (1,0.25)  -- (1,-0.25);
  \draw (2,0.25)  -- (2,-0.25);
  \draw (3,0.25)  -- (3,-0.25);
  \draw (4,0.25)  -- (4,-0.25);
  \draw (5,0.25)  -- (5,-0.25);
  \draw (6,0.25)  -- (6,-0.25);
  \draw (7,0.25)  -- (7,-0.25);
  \draw (8,0.25)  -- (8,-0.25);
  \draw (9,0.25)  -- (9,-0.25);
  \draw (10,0.25) -- (10,-0.25);
  \draw (11,0.25) -- (11,-0.25);
  \draw (12,0.25) -- (12,-0.25);
  \draw (13,0.25) -- (13,-0.25);
  \draw (14,0.25) -- (14,-0.25);
  \draw (15,0.25) -- (15,-0.25);
  \draw (16,0.25) -- (16,-0.25);
  \draw (17,0.25) -- (17,-0.25);
  \draw (18,0.25) -- (18,-0.25);
  \draw (19,0.25) -- (19,-0.25);
  \draw (20,0.25) -- (20,-0.25);
  \draw (21,0.25) -- (21,-0.25);
  \draw (22,0.25) -- (22,-0.25);
  \draw (23,0.25) -- (23,-0.25);
  \draw (24,0.25) -- (24,-0.25);
  \draw (25,0.25) -- (25,-0.25);

  \node at (0, -0.7)  {\scalebox{2.3}{0}};
  \node at (5, -0.7)  {\scalebox{2.3}{5}};
  \node at (10, -0.7) {\scalebox{2.3}{10}};
  \node at (15, -0.7) {\scalebox{2.3}{15}};
  \node at (20, -0.7) {\scalebox{2.3}{20}};
  \node at (25, -0.7) {\scalebox{2.3}{25}};

  \fill[black] (5, 0) circle (0.25);
  \fill[black] (20, 0) circle (0.25);

  \draw (0,  -1.2) -- (0,  -1.6) -- (10, -1.6) -- (10, -1.2);
  \draw (15, -1.2) -- (15, -1.6) -- (25, -1.6) -- (25, -1.2);

  \node at (5, -2.5)  {\scalebox{2.3}{$\delta^{5}_{5}$}};
  \node at (20, -2.5) {\scalebox{2.3}{$\delta^{5}_{20}$}};

  \draw[line width=3] (3.75, -0.25) -- (4.25,0.25);
  \draw[line width=3] (3.75, 0.25)  -- (4.25,-0.25);

  \draw[line width=3] (6.75, -0.25) -- (7.25,0.25);
  \draw[line width=3] (6.75, 0.25)  -- (7.25,-0.25);

  \draw[line width=3] (11.75, -0.25) -- (12.25,0.25);
  \draw[line width=3] (11.75, 0.25)  -- (12.25,-0.25);

  \node at (4,  1.5) {\scalebox{2.3}{$TP$}};
  \node at (7,  1.5) {\scalebox{2.3}{$FP$}};
  \node at (12, 1.5) {\scalebox{2.3}{$FP$}};
  \node at (20, 1.5) {\scalebox{2.3}{$FN$}};

  \draw (4, 1)  -- (4, 0.35);
  \draw (7, 1)  -- (7, 0.35);
  \draw (12, 1) -- (12, 0.35);
  \draw (20, 1) -- (20, 0.35);

\end{tikzpicture}
    }
    \caption{Change results classification in $TP$, $FP$ and $FN$. A dot represents a real change. A cross represents a detection. Shadowed in the neighborhood.}
    \label{figure:changes:classification}
  \end{figure}
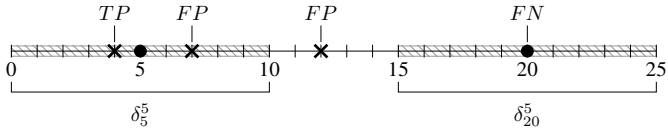

  To classify the detected changes as \emph{true positives} ($TP$), \emph{false positives} ($FP$) or \emph{false negatives} ($FN$), we use a threshold $\varepsilon$, that represents the error tolerance of the quality measures, and a neighborhood $\delta^{\varepsilon}_{i}$, defined as the interval between $i-\varepsilon$ and $i+\varepsilon$.
  Let a change happen at instant $i$.
  This change is classified as a $TP$ only when it is detected in $\delta^{\varepsilon}_{i}$.
  When no change is detected in $\delta^{\varepsilon}_{i}$ it is classified as $FN$.
  Finally, all changes detected in $\delta^{\varepsilon}_i$ where a previous change has been already detected is classified as a $FP$, as well as the ones detected outside any $\delta^{\varepsilon}$.
  \figurename~\ref{figure:changes:classification} shows an example with two real changes ($d_5$, at instant \num{5}, and $d_{20}$, at instant \num{20}), and three detections, at instants \num{4} ($c_4$), \num{7} ($c_7$) and \num{12} ($c_{12}$), using a $\varepsilon=5$.
  In this example, $c_4$ is classified as a $TP$, because it lies in the neighborhood of $d_5$; $c_7$ is classified as a $FP$, because, despite being in the neighborhood of $d_5$, another change has been detected previously; $c_{12}$ is classified too as a $FP$, in this case for being detected outside any neighborhood $\delta^5$; and finally, $d_{20}$ is classified as a $FN$ since no change is detected in its neighborhood.

  Both the algorithm and the data used for the tests are available online\footnote{tec.citius.usc.es/concept-drift-api/swagger-ui.html}.

  \subsection{Validation Data}
  \label{section:validation:data}
  To test the proposed approach, synthetic logs have been generated using the metodology, models and change patterns defined in \cite{DBLP:journals/tkde/MaaradjiDRO17}.
  This is the most accepted methodology for generating datasets when validating sudden concept drift detection algorithms in process mining.
  The generated dataset contains \num{68} logs: \num{17} with \num{2500} traces, \num{17} with \num{5000}, \num{17} with \num{7500} and \num{17} with \num{10000}.
  The original dataset from \cite{DBLP:journals/tkde/MaaradjiDRO17} contains 4 more logs (one for each of the sizes), but they have been discarded because its drifts (changing the frequency of the branches in a choice construct) are not control-flow drifts, but behavioural ones.

  \begin{table}[t]
    \centering
    \caption{Simple change patterns from \cite{DBLP:journals/dke/WeberRR08} applied to the original model.}
    \label{table:experimentation:change-patterns}
\tabulinesep=2pt
\begin{tabu}{cXc}
  \textbf{Code} & \centering{\textbf{Change pattern}}          & \textbf{Class} \\ \toprule
  \texttt{cm}   & Move fragment into/out of conditional branch & \texttt{I}     \\
  \texttt{cp}   & Duplicate fragment                           & \texttt{I}     \\
  \texttt{pm}   & Move fragment into/out of parallel branch    & \texttt{I}     \\
  \texttt{re}   & Add/remove fragment                          & \texttt{I}     \\
  \texttt{rp}   & Substitute fragment                          & \texttt{I}     \\
  \texttt{sw}   & Swap two fragments                           & \texttt{I}     \\ \tabucline[.4pt on 1pt black!60 off 4pt]{-}
  \texttt{cb}   & Make fragment skippable/non-skippable        & \texttt{O}     \\
  \texttt{lp}   & Make fragments loopable/non-loopable         & \texttt{O}     \\ \tabucline[.4pt on 1pt black!60 off 4pt]{-}
  \texttt{cd}   & Synchronize two fragments                    & \texttt{R}     \\
  \texttt{cf}   & Make two fragments conditional/sequential    & \texttt{R}     \\
  \texttt{pl}   & Make two fragments parallel/sequential       & \texttt{R}     \\ \bottomrule
\end{tabu}

  \end{table}

  \begin{figure}[t]
    \centering
    \resizebox {\linewidth} {!} {
      \input{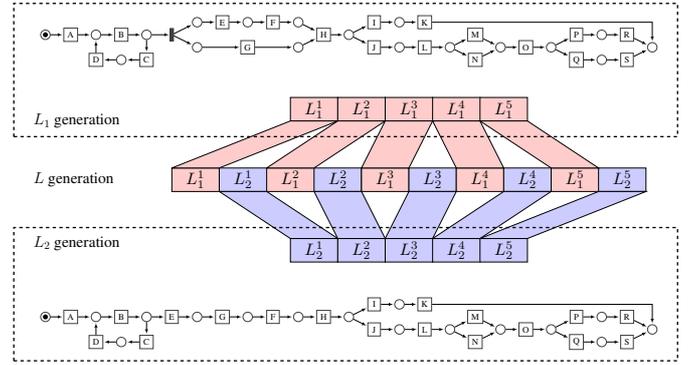}
    }
    \caption{Log generation example.}
    \label{figure:validation:log-generation}
  \end{figure}

  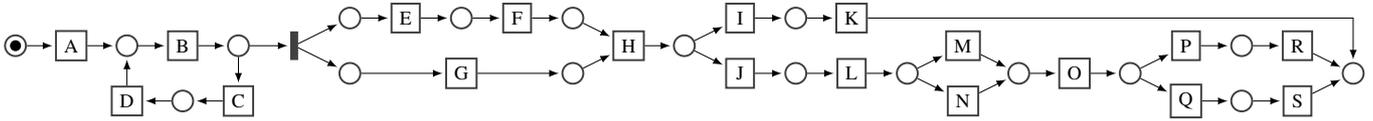
\begin{figure*}[t]
    \centering
    \resizebox {\textwidth} {!} {

\begin{tikzpicture}[node distance = 8mm, >=latex, bend angle=45, auto]
  \tikzstyle{place} = [circle, thick, draw=black!75, minimum size=3mm, inner sep=0pt]
  \tikzstyle{transition} = [rectangle, thick, draw=black!75, minimum size=4mm]
  \tikzstyle{silent} = [rectangle, thin, draw=black!75, fill=black!75, minimum width=1mm, minimum height=4mm, inner sep=0pt]

  \node [place,tokens=1] (p0)  []                                       {};
  \node [place]          (p1)  [right of=p0,  xshift= 8mm]              {};
  \node [place]          (p2)  [right of=p1,  xshift= 8mm]              {};
  \node [place]          (p3)  [below of=p2,  xshift=-8mm]              {};
  \node [place]          (p4)  [right of=p2,  xshift= 8mm, yshift= 4mm] {};
  \node [place]          (p5)  [right of=p2,  xshift= 8mm, yshift=-4mm] {};
  \node [place]          (p6)  [right of=p4,  xshift= 8mm]              {};
  \node [place]          (p7)  [right of=p6,  xshift= 8mm]              {};
  \node [place]          (p8)  [right of=p5,  xshift=24mm]              {};
  \node [place]          (p9)  [right of=p8,  xshift= 8mm, yshift= 4mm] {};
  \node [place]          (p10) [right of=p9,  xshift= 8mm, yshift= 4mm] {};
  \node [place]          (p11) [right of=p9,  xshift= 8mm, yshift=-4mm] {};
  \node [place]          (p12) [right of=p11, xshift= 8mm]              {};
  \node [place]          (p13) [right of=p12, xshift= 8mm]              {};
  \node [place]          (p14) [right of=p13, xshift= 8mm]              {};
  \node [place]          (p15) [right of=p14, xshift= 8mm, yshift= 4mm] {};
  \node [place]          (p16) [right of=p14, xshift= 8mm, yshift=-4mm] {};
  \node [place]          (p17) [right of=p15, xshift= 8mm, yshift=-4mm] {};

  \node [transition] (T1)  [right of=p0]               {\footnotesize A}
    edge [pre]   (p0)
    edge [post]  (p1);
  \node [transition] (T2)  [right of=p1]               {\footnotesize B}
    edge [pre]   (p1)
    edge [post]  (p2);
  \node [transition] (T3)  [below of=p2]               {\footnotesize C}
    edge [pre]   (p2)
    edge [post]  (p3);
  \node [transition] (T4)  [below of=p1]               {\footnotesize D}
    edge [pre]   (p3)
    edge [post]  (p1);
  \node [transition] (T5)  [right of=p4]               {\footnotesize E}
    edge [pre]   (p4)
    edge [post]  (p6);
  \node [transition] (T6)  [right of=p6]               {\footnotesize F}
    edge [pre]   (p6)
    edge [post]  (p7);
  \node [transition] (T7)  [right of=p5, xshift=8mm]   {\footnotesize G}
    edge [pre]   (p5)
    edge [post]  (p8);
  \node [transition] (T8)  [left  of=p9]               {\footnotesize H}
    edge [pre]   (p7)
    edge [pre]   (p8)
    edge [post]  (p9);
  \node [transition] (T9)  [left  of=p10]              {\footnotesize I}
    edge [pre]   (p9)
    edge [post]  (p10);
  \node [transition] (T10) [left  of=p11]              {\footnotesize J}
    edge [pre]   (p9)
    edge [post]  (p11);
  \node [transition] (T11) [right of=p10]              {\footnotesize K}
    edge [pre]   (p10)
    edge [post, to path={-| (\tikztotarget)}]  (p17);
  \node [transition] (T12) [right of=p11]              {\footnotesize L}
    edge [pre]   (p11)
    edge [post]  (p12);
  \node [transition] (T13) [right of=p12, yshift=4mm]  {\footnotesize M}
    edge [pre]   (p12)
    edge [post]  (p13);
  \node [transition] (T14) [right of=p12, yshift=-4mm] {\footnotesize N}
    edge [pre]   (p12)
    edge [post]  (p13);
  \node [transition] (T15) [right of=p13]              {\footnotesize O}
    edge [pre]   (p13)
    edge [post]  (p14);
  \node [transition] (T16) [left  of=p15]              {\footnotesize P}
    edge [pre]   (p14)
    edge [post]  (p15);
  \node [transition] (T17) [left  of=p16]              {\footnotesize Q}
    edge [pre]   (p14)
    edge [post]  (p16);
  \node [transition] (T18) [right of=p15]              {\footnotesize R}
    edge [pre]   (p15)
    edge [post]  (p17);
  \node [transition] (T19) [right of=p16]              {\footnotesize S}
    edge [pre]   (p16)
    edge [post]  (p17);

  \node [silent] (S1) [right of=p2] {}
    edge [pre]   (p2)
    edge [post]  (p4)
    edge [post]  (p5);
\end{tikzpicture}
    }
    \caption{Petri net for the original model\cite{DBLP:journals/tkde/MaaradjiDRO17} used to generate logs. Activity names are shortened for better understandability.}
    \label{figure:validation:model}
  \end{figure*}

  The original model used to generate the logs corresponds to a loan application process.
  The corresponding Petri net is depicted in \figurename~\ref{figure:validation:model}.
  To generate the \num{17} modified models, \num{11} simple change patterns from \cite{DBLP:journals/dke/WeberRR08} are applied to the original process.
  The applied patterns are collected in \tablename~\ref{table:experimentation:change-patterns}.
  These changes can imply an insertion (labeled as $I$), an optionalization (labeled as $O$) or a resequentialization (labeled as $R$).
  For each simple change pattern a different model is generated.
  The remaining 6 models are generated by applying a combination of simple change patterns, picking one change from each of the previously named classes.

  Once all the models are available, the logs are generated simulating executions of those processes.
  The original log is then combined with the modified ones to generate logs with drifts.
  The final log is composed joining alternatively sublogs from both the original model and the modified ones.
  Each drifting log presents a change every \SI{10}{\percent} of its final size.
  A log generation example is represented in \figurename~\ref{figure:validation:log-generation}.
  Two logs ($L_1$ and $L_2$) with different models are split in 5 sublogs with equal sizes ($L^1_1$ to $L^5_1$ and $L^1_2$ to $L^5_2$).
  This sublogs are combined alternatively into a log $L$, with size $|L_1| + |L_2|$.

\subsection{Impact of Discovery Algorithm and Fitness and Precision Metrics in Detection}\label{section:validation:impact-fitness-precision}
  In order to check the impact of the discovery algorithm and the conformance metrics in terms of both $F_{score}$, $\Delta$ and computation time, different configurations have been tested:

  \begin{enumerate}
    \item Discovery algorithms: Inductive Miner ($IM$)\cite{DBLP:conf/apn/LeemansFA13} and Heuristics Miner ($HM$)\cite{DBLP:conf/cidm/WeijtersR11}, which are two of the most used methods for discovering models from event logs.
    No algorithm based on evolutionary computation has been selected because it would increase the computational complexity significantly.
    \item Fitness metrics: Alignment Based Fitness ($AF$)\cite{DBLP:journals/is/RozinatA08}, Negative Event Recall ($NR$)\cite{DBLP:journals/tkde/BrouckeWVB14} and the percentage of completely replayable traces ($RT$).
    \item Precision metrics,: Advanced Behavioural Appropriateness ($ABA$)\cite{DBLP:journals/is/RozinatA08}, Negative Event Precision ($NP$)\cite{DBLP:journals/tkde/BrouckeWVB14} and precision change assessment ($PC$).
  \end{enumerate}

  \begin{table}[t]
    \centering
    \caption{Mean $\Delta$ and $F_{score}$ for every tested configuration}
    \label{table:validation:results:our}
\tabulinesep=2pt

\begin{tabu}{lllSS}
      &             &            & $F_{score}$      & $\Delta$                  \\\toprule
      &             & $\rho=ABA$ &                              0.9028 &                              4.0915 \\
      & $\gamma=AF$ & $\rho=NP$  &                              0.7985 &                             43.6243 \\
      &             & $\rho=PC$  & {\cellcolor{black!30!white}} 0.9969 &                              3.6471 \\\cmidrule{2-5}
      &             & $\rho=ABA$ &                              0.9028 &                              4.0915 \\
$IM$  & $\gamma=NR$ & $\rho=NP$  &                              0.7864 &                             41.5318 \\
      &             & $\rho=PC$  & {\cellcolor{black!30!white}} 0.9969 & {\cellcolor{black!15!white}} 3.5948 \\\cmidrule{2-5}
      &             & $\rho=ABA$ &                              0.9425 &                              4.0663 \\
      & $\gamma=RT$ & $\rho=NP$  &                              0.7955 &                             41.7028 \\
      &             & $\rho=PC$  & {\cellcolor{black!30!white}} 0.9969 & {\cellcolor{black!15!white}} 3.5948 \\\midrule
      &             & $\rho=ABA$ &                              0.9327 &                              7.4608 \\
      & $\gamma=AF$ & $\rho=NP$  &                              0.7365 &                             36.1014 \\
      &             & $\rho=PC$  & {\cellcolor{black!15!white}} 0.9789 &                              4.4412 \\\cmidrule{2-5}
      &             & $\rho=ABA$ &                              0.9402 &                             10.3758 \\
$HM$  & $\gamma=NR$ & $\rho=NP$  &                              0.7427 &                             36.8831 \\
      &             & $\rho=PC$  &                              0.9750 &                              5.3828 \\\cmidrule{2-5}
      &             & $\rho=ABA$ &                              0.4724 &                              6.0812 \\
      & $\gamma=RT$ & $\rho=NP$  &                              0.7025 &                             73.5175 \\
      &             & $\rho=PC$  &                              0.7176 & {\cellcolor{black!30!white}} 2.9829 \\\bottomrule\vspace{-1mm}
\textit{$\scriptstyle IM$}  & \multicolumn{4}{l}{\scriptsize Inductive Miner} \\\vspace{-1mm}
\textit{$\scriptstyle HM$}  & \multicolumn{4}{l}{\scriptsize Heuristics Miner} \\\vspace{-1mm}
\textit{$\scriptstyle AF$}  & \multicolumn{4}{l}{\scriptsize Alignment Based Fitness} \\\vspace{-1mm}
\textit{$\scriptstyle NR$}  & \multicolumn{4}{l}{\scriptsize Negative Event Recall} \\\vspace{-1mm}
\textit{$\scriptstyle RT$}  & \multicolumn{4}{l}{\scriptsize Percentage of Replayable Traces} \\\vspace{-1mm}
\textit{$\scriptstyle ABA$} & \multicolumn{4}{l}{\scriptsize Advanced Behavioural Appropriateness} \\\vspace{-1mm}
\textit{$\scriptstyle NP$}  & \multicolumn{4}{l}{\scriptsize Negative Event Precision} \\\vspace{-1mm}
\textit{$\scriptstyle PC$}  & \multicolumn{4}{l}{\scriptsize Precision Change Assessment} \\\bottomrule
\end{tabu}

  \end{table}

  \textbf{Impact on $F_{score}$ and $\Delta$.~}
  \tablename~\ref{table:validation:results:our} shows the mean $F_{score}$ and $\Delta$ values for each tested combination of discovery algorithm, fitness anprecision metrics using a window size of \num{100} traces and logs with \num{2500} traces.
  Best results are shadowed in dark and second best results in light grey.
  In order to compute the $F_{score}$, the error tolerance $\varepsilon$ has been set to a \SI{5}{\percent} of the log size, that is, half of the trace counbetween real changes.

  As can be seen in these results, the parameter with greater impact is the precision metric, being $NP$ precision the one that presents the worst detection resultin terms of both $F_{score}$ and $\Delta$.
  This metric has a greater sensitivity, so small changes in the frequency of activities in the processed window can dramatically change the precision, and thus bmistaken with structural changes.
  $ABA$ precision obtains good results in most of cases, but does not finish in some logs, causing some false negatives that affect the final results.
  Best results are archived using $PC$ precision, as it is insensitive to the activity frequency, counting only if relations appear at least once in the analyzewindow.

  Fitness has a smaller impact on results, and all tested metrics obtain very similar results, specially in the case of the $IM$, which ensures a complete fitnessno matter which metric is used to compute it.
  Only the combination of $RT$ fitness with $HM$ algorithm gets worse results since $HM$ does not ensure a complete trace replayability.

\begin{figure}[t]
    \centering
    \resizebox {\linewidth} {!} {

\begin{tikzpicture}
  \pgfplotsset{every y tick label/.append style={font=\scriptsize}}
  \pgfplotsset{every y tick label/.append style={font=\footnotesize}}
  \pgfplotsset{every node near coord/.append style={font=\scriptsize}}

  \begin{scope}
    \begin{axis}[
      width=.5\linewidth,
      height=.7\linewidth,
      ybar,
      ymin=0,
      ymax=650,
      xmin=0.5,
      xmax=3.5,
      ylabel={Execution time (\si{\ms})},
      xtick=data,
      nodes near coords,
      title={Fitness},
      bar width=6mm,
      xtick=\empty,
      ytick={100, 200, 300, 400, 500, 600},
      axis y line*=left,
      axis x line*=bottom,
      axis x line =bottom,
      x axis line style=-,
      y axis line style={draw=none},
      title style={font=\footnotesize, at={(0.5, -0.05)},anchor=north},
      bar shift = 0,
      label style={font=\footnotesize}
    ]
      \addplot[black,fill=orange!40!white, postaction={pattern = crosshatch dots,  pattern color=orange!60!black}] coordinates {(1, 237.8)};
      \addplot[black,fill=orange!40!white, postaction={pattern = north west lines, pattern color=orange!60!black}] coordinates {(2, 85.76)};
      \addplot[black,fill=orange!40!white, postaction={pattern = north east lines, pattern color=orange!60!black}] coordinates {(3, 65.68)};
    \end{axis}
  \end{scope}

  \begin{scope}[xshift = .4\linewidth]
    \begin{axis}[
      width=.5\linewidth,
      height=.7\linewidth,
      ybar,
      area legend,
      legend entries={$AF$,$NR$,$RT$,$ABA$,$NP$,$PC$},
      legend style={font=\scriptsize, at={(0.805, 1.05)}, anchor=south east, draw=none, /tikz/every even column/.append style={column sep=5mm}},
      legend columns = 3,
      ymin=0,
      ymax=650,
      xmin=0.5,
      xmax=3.5,
      nodes near coords,
      title={Precision},
      bar width=6mm,
      xtick=\empty,
      ytick={100, 200, 300, 400, 500, 600},
      axis y line*=right,
      axis x line*=bottom,
      axis x line =bottom,
      x axis line style=-,
      y axis line style={draw=none},
      title style={font=\footnotesize, at={(0.5, -0.05)},anchor=north},
      bar shift = 0,
      label style={font=\footnotesize}
    ]
      \addplot[black, fill=orange!40!white, postaction={pattern = crosshatch dots,  pattern color=orange!60!black}] coordinates {(-1, 0)};
      \addplot[black, fill=orange!40!white, postaction={pattern = north west lines, pattern color=orange!60!black}] coordinates {(-2, 0)};
      \addplot[black, fill=orange!40!white, postaction={pattern = north east lines, pattern color=orange!60!black}] coordinates {(-3, 0)};
      \addplot[black, fill=blue!40!white,   postaction={pattern = crosshatch dots,  pattern color=blue!60!black}]   coordinates {(1, 611.36)};
      \addplot[black, fill=blue!40!white,   postaction={pattern = north west lines, pattern color=blue!60!black}]   coordinates {(2, 490.88)};
      \addplot[black, fill=blue!40!white,   postaction={pattern = north east lines, pattern color=blue!60!black}]   coordinates {(3,   7.52)};
    \end{axis}
  \end{scope}
\end{tikzpicture}
    }
    \caption{Average execution time for each one of the tested metrics.}
    \label{figure:validation:timing}
  \end{figure}
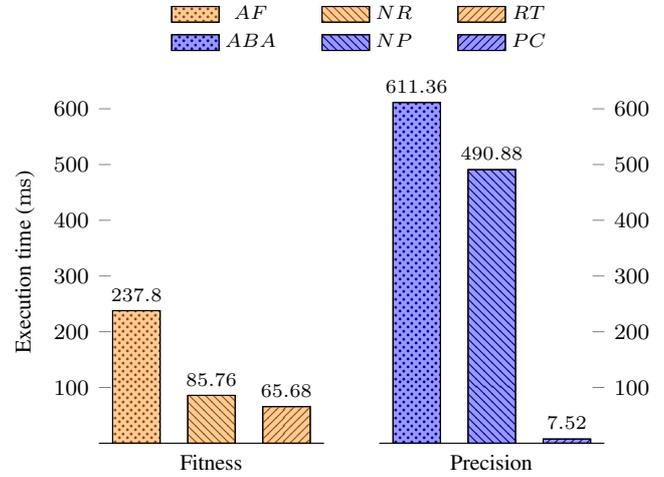

  \textbf{Impact on computation time.~}
  \figurename~\ref{figure:validation:timing} summarizes the average time required for processing a window of \num{100} traces by each metric.
  Test were run on an Intel(R) Core(TM) i7 860 and \SI{16}{\gibi\byte} of RAM.
  As we can see, $PC$ precision is sensitively faster than the rest of the precision metrics.
  In the case of fitness, the difference is much lower, being $RT$ the faster one, closely followed by $NR$.

  On the basis of the above, the experiments in the following sections have been performed using $IM$ algorithm, $RT$ fitness and $PC$ precision.

\subsection{Impact of the Window Size}
  \begin{figure}[t]
    \centering
    \resizebox {\linewidth} {!} {

\begin{tikzpicture}
  \begin{axis}[
    axis x line*=none,
    axis x line=none,
    x axis line style=-,
    axis y line=none,
    axis y line*=none,
    xmin=0,
    xmax=500,
    ymax=1.05,
    ymin=0
  ]
    \addplot[name path = min, mark=none, white!0!white] coordinates{(20, 0) (20, 70)};
    \addplot[name path = max, mark=none, white!0!white] coordinates{(105, 0) (105, 70)};
    \addplot fill between[of = min and max, every even segment/.style = {pattern=crosshatch, pattern color=green!30!white}];
  \end{axis}

  \begin{axis}[
    ylabel={$F_{score}$},
    axis x line*=bottom,
    axis x line=bottom,
    x axis line style=-,
    axis y line=left,
    axis y line*=left,
    xmin=0,
    xmax=500,
    ymax=1.05,
    ymin=0,
    xtick={10,25,50,100,150,200,250,300,400,500},
    xticklabels={10,,50,100,150,200,250,300,400,500},
    legend style={at={(0.5,1.0)}, anchor=south east, draw=none, /tikz/every even column/.append style={column sep=0.5cm}},
    legend columns=-1,
    legend entries={$F_{score}$},
  ]
    \addplot[red!60, mark=*] coordinates {
      (10,  0.5651)
      (25,  0.9569)
      (50,  0.9938)
      (100, 0.9969)
      (150, 0.6907)
      (200, 0.3842)
      (250, 0.1861)
      (300, 0.0214)
      (400, 0.0214)
      (500, 0.0214)
    };
  \end{axis}

  \begin{axis}[
    ylabel={$\Delta$},
    axis x line*= none,
    axis x line = bottom,
    x axis line style=-,
    axis y line=right,
    axis y line* = right,
    xmin=0,
    xmax=500,
    ymin=0,
    ymax=70,
    legend style={at={(0.5,1.0)}, anchor=south west, draw=none, /tikz/every even column/.append style={column sep=0.5cm}},
    legend columns=-1,
    xtick={10,25,50,100,150,200,250,300,400,500},
    xticklabels=\empty,
    legend entries={$\Delta$},
  ]
    \addplot[blue!60, mark=x] coordinates {
      (10,   8.0498)
      (25,   3.5033)
      (50,   3.3856)
      (100,  3.5948)
      (150,  4.7935)
      (200, 48.0389)
      (250, 62.5333)
      (300, 50.5000)
      (400, 50.5000)
      (500, 50.5000)
    };
  \end{axis}
\end{tikzpicture}
    }
    \caption{Mean $F_{score}$ (higger is better) and $\Delta$ (lower is better) evolution using different window sizes. Shadowed in green is the optimal region.}
    \label{figure:validation:results:window-size}
  \end{figure}
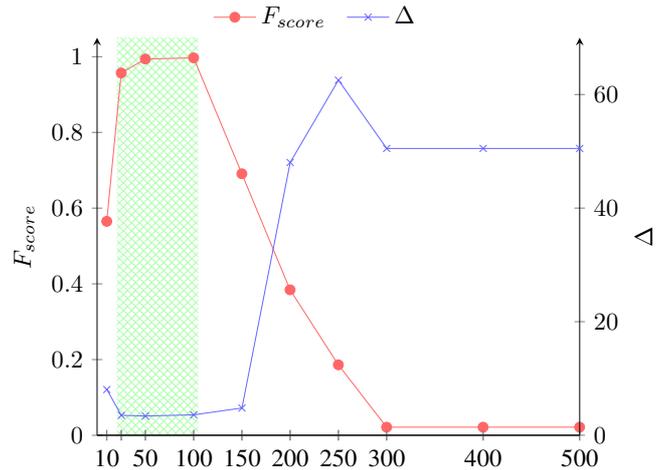

  \figurename~\ref{figure:validation:results:window-size} shows the impact of the window size in both mean $F_{score}$ and mean $\Delta$ for logs of \num{2500} traces.
  The optimal window size for both $F_{score}$ and $\Delta$ ranges between \num{25} and \num{100} (this area is boxed in green).
  Reduced $\Delta$, but with low $F_{score}$, are obtained for windows with at most \num{25} traces.
  In this case, \proposedname~algorithm identifies many false positives because the reference window does not include enough behavior to discover the model.
  \proposedname~also gets low $F_{score}$ for windows with more than \num{100} traces because false negatives are identified as the window contains behavior of both before and after the change.
  In the experiments of the following sections, we will use a window of \num{100} traces, which is the value that produces the better $F_{score}$ without increasing the $\Delta$.

\subsection{Comparison with Other Process Drift Detection Algorithms}
  \begin{table*}[t]
    \centering
    \caption{Mean $F_{score}$ and $\Delta$ values for each algorithm using logs with \num{2500} traces.}
    \label{table:validation:results:comparative:2500}
\tabulinesep=2pt
\begin{tabu}{l p{5mm} X[c]X[c] p{1mm} X[c]X[c] p{1mm} X[c]X[c] p{1mm} X[c]X[c]}
  \multirow{2}{*}{\textbf{\textit{\texttt{Log}}}} &&\multicolumn{2}{c}{\textit{\proposedname}}&&\multicolumn{2}{c}{\textit{TPCDD}} && \multicolumn{2}{c}{\textit{PD-T}}  && \multicolumn{2}{c}{\textit{PD-E}}  \\
                        \cmidrule{3-4}\cmidrule{6-7}\cmidrule{9-10}\cmidrule{12-13}
                        &&$F_{score}$                          &$\Delta$&&$F_{score}$                          &$\Delta$ &&$F_{score}$                          &$\Delta$ &&$F_{score}$                          &$\Delta$   \\\toprule
  \texttt{\textbf{cb}}  && {\cellcolor{black!20!white}} 1.0000 & 6.2222 && {\cellcolor{black!20!white}} 1.0000 &  5.5556 &&                              0.2000 & 90.0000 && {\cellcolor{black!20!white}} 1.0000 & 78.4444   \\
  \texttt{\textbf{cd}}  && {\cellcolor{black!20!white}} 1.0000 & 1.7778 && {\cellcolor{black!20!white}} 1.0000 &  0.7778 &&                              0.0000 &  {---}  && {\cellcolor{black!20!white}} 1.0000 & 42.4444   \\
  \texttt{\textbf{cf}}  && {\cellcolor{black!20!white}} 1.0000 & 3.8889 && {\cellcolor{black!20!white}} 1.0000 &  1.4444 && {\cellcolor{black!20!white}} 1.0000 & 40.0000 && {\cellcolor{black!20!white}} 1.0000 & 49.7778   \\
  \texttt{\textbf{cm}}  && {\cellcolor{black!20!white}} 1.0000 & 5.8889 && {\cellcolor{black!20!white}} 1.0000 & 11.5556 &&                              0.7143 & 66.0000 &&                              0.9412 & 95.1250   \\
  \texttt{\textbf{cp}}  && {\cellcolor{black!20!white}} 1.0000 & 2.8889 && {\cellcolor{black!20!white}} 1.0000 &  1.2222 && {\cellcolor{black!20!white}} 1.0000 & 45.2222 && {\cellcolor{black!20!white}} 1.0000 & 39.0000   \\
  \texttt{\textbf{lp}}  && {\cellcolor{black!20!white}} 1.0000 & 3.6667 &&                              0.9474 & 10.0000 && {\cellcolor{black!20!white}} 1.0000 & 52.7778 &&                              0.9412 & 36.5000   \\
  \texttt{\textbf{sw}}  && {\cellcolor{black!20!white}} 1.0000 & 2.8889 && {\cellcolor{black!20!white}} 1.0000 &  1.0000 && {\cellcolor{black!20!white}} 1.0000 & 43.5556 &&                              0.9412 & 38.1250   \\
  \texttt{\textbf{pl}}  && {\cellcolor{black!20!white}} 1.0000 & 1.7778 && {\cellcolor{black!20!white}} 1.0000 &  1.3333 &&                              0.9412 & 36.0000 &&                              0.9412 & 51.3750   \\
  \texttt{\textbf{pm}}  && {\cellcolor{black!20!white}} 1.0000 & 3.2222 && {\cellcolor{black!20!white}} 1.0000 &  3.3333 &&                              0.8000 & 49.6667 && {\cellcolor{black!20!white}} 1.0000 & 48.4444   \\
  \texttt{\textbf{re}}  && {\cellcolor{black!20!white}} 1.0000 & 2.0000 && {\cellcolor{black!20!white}} 1.0000 &  1.1111 && {\cellcolor{black!20!white}} 1.0000 & 19.8889 &&                              0.9412 & 49.8750   \\
  \texttt{\textbf{rp}}  && {\cellcolor{black!20!white}} 1.0000 & 3.0000 && {\cellcolor{black!20!white}} 1.0000 &  1.2222 && {\cellcolor{black!20!white}} 1.0000 & 45.2222 && {\cellcolor{black!20!white}} 1.0000 & 65.0000   \\
  \texttt{\textbf{IOR}} && {\cellcolor{black!20!white}} 1.0000 & 2.6667 && {\cellcolor{black!20!white}} 1.0000 &  2.0000 && {\cellcolor{black!20!white}} 1.0000 & 38.7778 && {\cellcolor{black!20!white}} 1.0000 & 63.3333   \\
  \texttt{\textbf{IRO}} && {\cellcolor{black!20!white}} 1.0000 & 6.2222 && {\cellcolor{black!20!white}} 1.0000 &  2.7778 && {\cellcolor{black!20!white}} 1.0000 & 52.3333 && {\cellcolor{black!20!white}} 1.0000 & 43.0000   \\
  \texttt{\textbf{OIR}} &&                              0.9474 & 5.7778 && {\cellcolor{black!20!white}} 1.0000 &  0.6667 &&                              0.6154 & 29.5000 &&                              0.9412 & 15.7500   \\
  \texttt{\textbf{ORI}} && {\cellcolor{black!20!white}} 1.0000 & 2.5556 && {\cellcolor{black!20!white}} 1.0000 &  0.6667 && {\cellcolor{black!20!white}} 1.0000 & 50.5556 && {\cellcolor{black!20!white}} 1.0000 & 35.0000   \\
  \texttt{\textbf{RIO}} && {\cellcolor{black!20!white}} 1.0000 & 4.6667 && {\cellcolor{black!20!white}} 1.0000 & 10.8889 &&                              0.9412 & 51.1250 &&                              0.9412 & 53.8750   \\
  \texttt{\textbf{ROI}} && {\cellcolor{black!20!white}} 1.0000 & 2.0000 && {\cellcolor{black!20!white}} 1.0000 &  0.7778 && {\cellcolor{black!20!white}} 1.0000 & 32.3333 &&                              0.9412 & 40.1250   \\\bottomrule
  \texttt{\textbf{AVG}} && {\cellcolor{black!20!white}} 0.9969 & 3.5948 && {\cellcolor{black!20!white}} 0.9969 &  3.3137 &&                              0.8360 & 46.4349 &&                              0.9723 & 49.7173
\end{tabu}

  \end{table*}

  \begin{table*}[t]
    \centering
    \caption{Mean $F_{score}$ and $\Delta$ values for each algorithm using logs with \num{5000} traces.}
    \label{table:validation:results:comparative:5000}
\tabulinesep=2pt
\begin{tabu}{l p{5mm} X[c]X[c] p{1mm} X[c]X[c] p{1mm} X[c]X[c] p{1mm} X[c]X[c]}
\multirow{2}{*}{\textbf{\textit{\texttt{Log}}}} &&\multicolumn{2}{c}{\textit{\proposedname}}&&\multicolumn{2}{c}{\textit{TPCDD}} && \multicolumn{2}{c}{\textit{PD-T}}  && \multicolumn{2}{c}{\textit{PD-E}}  \\
                        \cmidrule{3-4}\cmidrule{6-7}\cmidrule{9-10}\cmidrule{12-13}
                        &&  $F_{score}$                        &$\Delta$&&$F_{score}$&$\Delta$&& $F_{score}$                         & $\Delta$&& $F_{score}$                         & $\Delta$ \\\toprule
  \texttt{\textbf{cb}}  && {\cellcolor{black!20!white}} 1.0000 & 6.7778 && 0.9000    & 6.5556 &&                              0.7143 & 57.4000 && {\cellcolor{black!20!white}} 1.0000 &  99.6667 \\
  \texttt{\textbf{cd}}  && {\cellcolor{black!20!white}} 1.0000 & 1.2222 && 0.9474    & 0.7778 &&                              0.0000 &  {---}  &&                              0.9412 &   2.5555 \\
  \texttt{\textbf{cf}}  && {\cellcolor{black!20!white}} 1.0000 & 3.3333 && 0.9474    & 9.2222 && {\cellcolor{black!20!white}} 1.0000 & 25.7777 && {\cellcolor{black!20!white}} 1.0000 &  50.1111 \\
  \texttt{\textbf{cm}}  && {\cellcolor{black!20!white}} 1.0000 & 7.7778 && 0.9474    & 6.1111 &&                              0.7143 & 51.8000 &&                              0.8750 & 117.5714 \\
  \texttt{\textbf{cp}}  && {\cellcolor{black!20!white}} 1.0000 & 3.5556 && 0.9000    & 5.4444 && {\cellcolor{black!20!white}} 1.0000 & 33.3333 && {\cellcolor{black!20!white}} 1.0000 &  66.2222 \\
  \texttt{\textbf{lp}}  && {\cellcolor{black!20!white}} 1.0000 & 2.0000 && 0.9000    & 4.1111 && {\cellcolor{black!20!white}} 1.0000 & 48.4444 && {\cellcolor{black!20!white}} 1.0000 &  45.3333 \\
  \texttt{\textbf{sw}}  && {\cellcolor{black!20!white}} 1.0000 & 2.7778 && 0.9474    & 0.6667 && {\cellcolor{black!20!white}} 1.0000 & 29.6667 && {\cellcolor{black!20!white}} 1.0000 &  21.3333 \\
  \texttt{\textbf{pl}}  && {\cellcolor{black!20!white}} 1.0000 & 1.2222 && 0.9474    & 0.7778 &&                              0.0000 &  {---}  && {\cellcolor{black!20!white}} 1.0000 &  51.2222 \\
  \texttt{\textbf{pm}}  && {\cellcolor{black!20!white}} 1.0000 & 3.6667 && 0.9474    & 3.4444 &&                              0.9412 & 37.0000 && {\cellcolor{black!20!white}} 1.0000 &  25.8889 \\
  \texttt{\textbf{re}}  && {\cellcolor{black!20!white}} 1.0000 & 2.0000 && 0.9474    & 1.5556 && {\cellcolor{black!20!white}} 1.0000 & 19.1111 &&                              0.9474 &  16.7778 \\
  \texttt{\textbf{rp}}  && {\cellcolor{black!20!white}} 1.0000 & 2.8889 && 0.9000    & 0.4444 && {\cellcolor{black!20!white}} 1.0000 & 28.2222 &&                              0.9412 &  48.7500 \\
  \texttt{\textbf{IOR}} && {\cellcolor{black!20!white}} 1.0000 & 2.2222 && 0.9474    & 2.7778 && {\cellcolor{black!20!white}} 1.0000 & 24.2222 && {\cellcolor{black!20!white}} 1.0000 &  52.0000 \\
  \texttt{\textbf{IRO}} && {\cellcolor{black!20!white}} 1.0000 & 5.4444 && 0.9000    & 1.7778 && {\cellcolor{black!20!white}} 1.0000 & 49.4444 &&                              0.9412 &  31.2500 \\
  \texttt{\textbf{OIR}} &&                              0.9474 & 7.7778 && 0.9000    & 1.0000 && {\cellcolor{black!20!white}} 1.0000 & 26.6667 &&                              0.9412 &   0.0000 \\
  \texttt{\textbf{ORI}} && {\cellcolor{black!20!white}} 1.0000 & 3.2222 && 0.9000    & 1.1111 && {\cellcolor{black!20!white}} 1.0000 & 36.1111 &&                              0.9412 &  22.7500 \\
  \texttt{\textbf{RIO}} && {\cellcolor{black!20!white}} 1.0000 & 5.2222 && 0.9474    & 1.7778 && {\cellcolor{black!20!white}} 1.0000 & 45.2222 && {\cellcolor{black!20!white}} 1.0000 &  59.7778 \\
  \texttt{\textbf{ROI}} && {\cellcolor{black!20!white}} 1.0000 & 2.0000 && 0.9474    & 0.4444 && {\cellcolor{black!20!white}} 1.0000 & 26.0000 &&                              0.9412 &   7.5000 \\\bottomrule
  \texttt{\textbf{AVG}} && {\cellcolor{black!20!white}} 0.9969 & 3.7124 && 0.9279    & 2.8235 &&                              0.8453 & 35.8948 &&                              0.9692
&  42.2771 \end{tabu}

  \end{table*}

  \begin{table*}[t]
    \centering
    \caption{Mean $F_{score}$ and $\Delta$ values for each algorithm using logs with \num{7500} traces.}
    \label{table:validation:results:comparative:7500}
\tabulinesep=2pt
\begin{tabu}{l p{5mm} X[c]X[c] p{1mm} X[c]X[c] p{1mm} X[c]X[c] p{1mm} X[c]X[c]}
\multirow{2}{*}{\textbf{\textit{\texttt{Log}}}} &&\multicolumn{2}{c}{\textit{\proposedname}}&&\multicolumn{2}{c}{\textit{TPCDD}} && \multicolumn{2}{c}{\textit{PD-T}}  && \multicolumn{2}{c}{\textit{PD-E}}  \\
                        \cmidrule{3-4}\cmidrule{6-7}\cmidrule{9-10}\cmidrule{12-13}
                        &&  $F_{score}$                        &$\Delta$&& $F_{score}$ &$\Delta$ &&  $F_{score}$                        & $\Delta$&& $F_{score}$&$\Delta$  \\\toprule
  \texttt{\textbf{cb}}  && {\cellcolor{black!20!white}} 1.0000 & 7.7778 &&      0.9474 & 16.6667 && {\cellcolor{black!20!white}} 1.0000 & 68.6667 && 0.9474     & 70.2222  \\
  \texttt{\textbf{cd}}  && {\cellcolor{black!20!white}} 1.0000 & 2.1111 &&      0.9474 &  1.0000 &&                              0.0000 &  {---}  && 0.9474     &  2.2222  \\
  \texttt{\textbf{cf}}  && {\cellcolor{black!20!white}} 1.0000 & 2.7778 &&      0.9474 &  1.5556 && {\cellcolor{black!20!white}} 1.0000 & 22.3333 && 0.9474     & 37.5556  \\
  \texttt{\textbf{cm}}  && {\cellcolor{black!20!white}} 1.0000 & 5.3333 &&      0.9474 &  4.4444 &&                              0.9412 & 83.8750 && 0.8889     & 78.6250  \\
  \texttt{\textbf{cp}}  && {\cellcolor{black!20!white}} 1.0000 & 3.2222 &&      0.9000 &  1.3333 && {\cellcolor{black!20!white}} 1.0000 & 33.2222 && 0.9474     & 26.5556  \\
  \texttt{\textbf{lp}}  && {\cellcolor{black!20!white}} 1.0000 & 1.4444 &&      0.7500 &  1.2222 && {\cellcolor{black!20!white}} 1.0000 & 52.8889 && 0.7200     & 34.6667  \\
  \texttt{\textbf{sw}}  && {\cellcolor{black!20!white}} 1.0000 & 2.6667 &&      0.8571 &  0.8889 && {\cellcolor{black!20!white}} 1.0000 & 32.4444 && 0.9000     & 16.1111  \\
  \texttt{\textbf{pl}}  && {\cellcolor{black!20!white}} 1.0000 & 2.1111 &&      0.9474 &  1.6667 &&                              0.0000 &  {---}  && 0.9474     & 30.8889  \\
  \texttt{\textbf{pm}}  && {\cellcolor{black!20!white}} 1.0000 & 3.1111 &&      0.9474 &  2.6667 && {\cellcolor{black!20!white}} 1.0000 & 43.4444 && 0.9474     & 14.6667  \\
  \texttt{\textbf{re}}  && {\cellcolor{black!20!white}} 1.0000 & 2.0000 &&      0.9474 &  0.5556 && {\cellcolor{black!20!white}} 1.0000 & 21.4444 && 0.9474     & 27.5556  \\
  \texttt{\textbf{rp}}  && {\cellcolor{black!20!white}} 1.0000 & 2.5556 &&      0.8571 & 30.6667 && {\cellcolor{black!20!white}} 1.0000 & 29.8889 && 0.9000     & 50.5556  \\
  \texttt{\textbf{IOR}} && {\cellcolor{black!20!white}} 1.0000 & 4.3333 &&      0.9474 & 10.7778 &&                              0.9000 & 34.5556 && 0.9474     & 53.2222  \\
  \texttt{\textbf{IRO}} && {\cellcolor{black!20!white}} 1.0000 & 4.2222 &&      0.9000 &  2.7778 && {\cellcolor{black!20!white}} 1.0000 & 53.4444 && 0.9474     & 14.3333  \\
  \texttt{\textbf{OIR}} &&                              0.7500 & 2.0000 &&      0.7826 &  0.3333 && {\cellcolor{black!20!white}} 1.0000 & 69.7778 && 0.9474     &  0.0000  \\
  \texttt{\textbf{ORI}} && {\cellcolor{black!20!white}} 1.0000 & 2.3333 &&      0.8571 &  0.8889 && {\cellcolor{black!20!white}} 1.0000 & 33.1111 && 0.9000     & 36.8889  \\
  \texttt{\textbf{RIO}} && {\cellcolor{black!20!white}} 1.0000 & 4.2222 &&      0.8571 &  2.4444 && {\cellcolor{black!20!white}} 1.0000 & 40.6667 && 0.9000     & 47.6667  \\
  \texttt{\textbf{ROI}} && {\cellcolor{black!20!white}} 1.0000 & 2.0000 &&      0.9474 &  1.0000 && {\cellcolor{black!20!white}} 1.0000 & 31.3333 && 0.9474     &  7.8888  \\\bottomrule
  \texttt{\textbf{AVG}} && {\cellcolor{black!20!white}} 0.9853 & 3.1895 &&      0.8993 &  4.7582 &&                              0.8730 & 43.4065 && 0.9194     & 32.3309
\end{tabu}

  \end{table*}

  \begin{table*}[t]
    \centering
    \caption{Mean $F_{score}$ and $\Delta$ values for each algorithm using logs with \num{10000} traces.}
    \label{table:validation:results:comparative:10000}
\tabulinesep=2pt
\begin{tabu}{l p{5mm} X[c]X[c] p{1mm} X[c]X[c] p{1mm} X[c]X[c] p{1mm} X[c]X[c]}
\multirow{2}{*}{\textbf{\textit{\texttt{Log}}}} &&\multicolumn{2}{c}{\textit{\proposedname}}&&\multicolumn{2}{c}{\textit{TPCDD}} && \multicolumn{2}{c}{\textit{PD-T}}  && \multicolumn{2}{c}{\textit{PD-E}}  \\
                        \cmidrule{3-4}\cmidrule{6-7}\cmidrule{9-10}\cmidrule{12-13}
                        && $F_{score}$                         &$\Delta$&& $F_{score}$&$\Delta$ &&$F_{score}$                          & $\Delta$&&$F_{score}$&$\Delta$ \\\toprule
  \texttt{\textbf{cb}}  && {\cellcolor{black!20!white}} 1.0000 & 9.0000 && 0.8571     & 18.2222 &&                              0.5000 & 41.6667 && 0.8571    & 84.4444 \\
  \texttt{\textbf{cd}}  && {\cellcolor{black!20!white}} 1.0000 & 1.5556 && 0.9474     &  1.0000 &&                              0.0000 &  {---}  && 0.9000    &  3.3333 \\
  \texttt{\textbf{cf}}  && {\cellcolor{black!20!white}} 1.0000 & 3.8889 && 0.9474     &  5.8889 && {\cellcolor{black!20!white}} 1.0000 & 29.6667 && 0.9474    & 45.8889 \\
  \texttt{\textbf{cm}}  && {\cellcolor{black!20!white}} 1.0000 & 8.5556 && 0.9000     &  7.6667 &&                              0.0000 &  {---}  && 0.9474    & 97.7778 \\
  \texttt{\textbf{cp}}  && {\cellcolor{black!20!white}} 1.0000 & 4.4444 && 0.9000     &  2.3333 && {\cellcolor{black!20!white}} 1.0000 & 33.5556 && 0.9000    & 30.7778 \\
  \texttt{\textbf{lp}}  && {\cellcolor{black!20!white}} 1.0000 & 2.3333 && 0.8182     &  8.1111 && {\cellcolor{black!20!white}} 1.0000 & 47.1111 && 0.8182    & 32.8889 \\
  \texttt{\textbf{sw}}  && {\cellcolor{black!20!white}} 1.0000 & 3.6667 && 0.9474     &  1.5556 && {\cellcolor{black!20!white}} 1.0000 & 33.4444 && 0.9000    &  4.6667 \\
  \texttt{\textbf{pl}}  && {\cellcolor{black!20!white}} 1.0000 & 1.5556 && 0.9474     &  1.1111 &&                              0.2000 & 84.0000 && 0.9474    & 30.7778 \\
  \texttt{\textbf{pm}}  && {\cellcolor{black!20!white}} 1.0000 & 3.6667 && 0.9474     &  2.1111 &&                              0.7500 & 31.3333 && 0.9474    &  9.0000 \\
  \texttt{\textbf{re}}  && {\cellcolor{black!20!white}} 1.0000 & 2.0000 && 0.9000     &  0.7778 && {\cellcolor{black!20!white}} 1.0000 & 17.7778 && 0.9000    & 23.0000 \\
  \texttt{\textbf{rp}}  && {\cellcolor{black!20!white}} 1.0000 & 3.7778 && 0.8182     &  1.3333 && {\cellcolor{black!20!white}} 1.0000 & 31.2222 && 0.8571    & 46.2222 \\
  \texttt{\textbf{IOR}} && {\cellcolor{black!20!white}} 1.0000 & 2.7778 && 0.9000     &  3.0000 && {\cellcolor{black!20!white}} 1.0000 & 27.7778 && 0.9474    & 59.5556 \\
  \texttt{\textbf{IRO}} && {\cellcolor{black!20!white}} 1.0000 & 7.6667 && 0.9474     &  2.1111 &&                              0.8750 & 50.1429 && 0.8571    & 18.0000 \\
  \texttt{\textbf{ORI}} && {\cellcolor{black!20!white}} 1.0000 & 3.8889 && 0.8182     &  1.1111 && {\cellcolor{black!20!white}} 1.0000 & 32.6667 && 0.9000    & 54.6666 \\
  \texttt{\textbf{OIR}} &&                              0.7826 & 2.0000 && 0.7200     &  0.2222 && {\cellcolor{black!20!white}} 1.0000 & 37.7778 && 0.9474    &  2.7778 \\
  \texttt{\textbf{RIO}} && {\cellcolor{black!20!white}} 1.0000 & 5.8889 && 0.7500     &  2.1111 &&                              0.8889 & 41.0000 && 0.8182    & 63.0000 \\
  \texttt{\textbf{ROI}} && {\cellcolor{black!20!white}} 1.0000 & 2.0000 && 0.9474     &  0.4444 && {\cellcolor{black!20!white}} 1.0000 & 20.6667 && 0.9000    &  8.4444 \\\bottomrule
  \texttt{\textbf{AVG}} && {\cellcolor{black!20!white}} 0.9872 & 4.0392 && 0.8831     &  3.4771 &&                              0.7773 & 37.8540 && 0.8995    & 36.1895
\end{tabu}

  \end{table*}

  In this section, \proposedname~algorithm is compared with
   \emph{Trace-Based ProDrift} (\textit{PD-T})\cite{DBLP:journals/tkde/MaaradjiDRO17},
   \emph{Event-Based ProDrift} (\textit{PD-E})\cite{DBLP:conf/er/OstovarMRHD16} and
   \textit{TPCDD} \cite{DBLP:conf/otm/ZhengW017},
  the three sudden process drift detection algorithms with best results in the state of the art.
  Specifically, we used the following configurations:

  \begin{enumerate}
    \item \textit{PD-T} with an \emph{adaptive window} and an initial size of \num{50};
    \item \textit{PD-E} with an \emph{adaptive window} and an initial size of \num{50}. \emph{Relation noise filter threshold} was set to \SI{0}{\percent} and \emph{sensitivity} to \emph{very high}, as the authors recommend for analyzing synthetic logs without noise;
    \item \textit{TPCDD} with minimum window size set to \num{100} and \emph{DBSCAN radius} to \num{10}.
  \end{enumerate}

  \tablename~\ref{table:validation:results:comparative:2500} to \tablename~\ref{table:validation:results:comparative:10000} show the detailed results.
  \proposedname~outperforms the rest of algorithms in terms of $F_{score}$ in all the cases, except in \texttt{OIR} log, getting always the best average value.
  Moreover, \proposedname~is also the second best in terms of delay, very close to \textit{TPCDD}, and being all the values in the same order of magnitude.

  For the \num{2500} trace logs, \proposedname~obtains a $F_{score}$ of \num{1.0} in all logs, except in \texttt{OIR}, where it returns a false positive.
  \textit{TPCDD} also obtains a perfect $F_{score}$, except in \texttt{lp}.
  In this case the error is due to a false negative (i.e., a change that remains undetected).
  Both \textit{PD-T} and \textit{PD-E} have much worse results, having \num{7} and \num{8} cases where they does not detect all changes.
  In fact, \textit{PD-T} even is unable to detect any change in \texttt{cd}.

  For logs with \num{5000}, \num{7500} and \num{10000} traces, \proposedname~gets similar results, and only do not have a perfect $F_{score}$ in \texttt{OIR}, again because of false positives.
  However, \textit{TPCDD} behaves much worse, being unable to get a perfect $F_{score}$ in any log.
  The same happens to \textit{PD-E} in logs with \num{7500} and \num{10000} traces, where it can not detect all changes correctly.
  \textit{PD-T} gets worse $F_{score}$ in most of the logs, and, in addition, is unable to detect any change in \texttt{cd} and \texttt{pl}, for logs with \num{5000} and \num{7500} traces, and in \texttt{cd} and \texttt{cm}, for logs with \num{10000} traces.

  In terms of $\Delta$, \proposedname~and \textit{TPCDD} obtain similar results for all the logs, being able to detect all changes always with less than 10 traces of delay.
  In comparison with \textit{PD-T} and \textit{PD-E}, \proposedname~$\Delta$ are always an order of magnitude below, thus being able to detect changes closer to the point where they really happened.

  \begin{figure}[t]
    \centering
    \resizebox {\linewidth} {!} {

\begin{tikzpicture}
  \pgfplotsset{every y tick label/.append style={font=\scriptsize}}
  \pgfplotsset{every x tick label/.append style={font=\footnotesize, rotate=60, anchor=east}}
  \pgfplotsset{every node near coord/.append style={font=\scriptsize}}

  \begin{scope}
    \begin{axis}[
      width=\linewidth,
      height=.5\linewidth,
      ybar,
      xmin = 0.5,
      xmax = 4.5,
      area legend,
      legend entries={\proposedname,\textit{TPCDD},\textit{PD-T},\textit{PD-E}},
      legend style={font=\footnotesize, at={(0.5,1.1)}, anchor=south, draw=none, /tikz/every even column/.append style={column sep=0.5cm}},
      legend columns = 4,
      ymin=0,
      ymax=1,
      xtick=\empty,
      ytick=\empty,
      nodes near coords,
      title={$F_{score}$},
      bar width= 8mm,
      axis y line =none,
      axis y line*=none,
      axis x line*=bottom,
      axis x line =bottom,
      x axis line style=-,
      title style={at={(0.5, -0.1)}, anchor=north},
      bar shift = 0,
    ]
      \addplot[black,fill=red!40!white, postaction={pattern = crosshatch dots, pattern color=red!60!black}] coordinates {(1, 0.9916)};
      \addplot[black,fill=blue!40!white, postaction={pattern = crosshatch, pattern color=blue!60!black}] coordinates {(2, 0.9268)};
      \addplot[black,fill=green!40!white, postaction={pattern = north west lines, pattern color=green!60!black}] coordinates {(3, 0.8329)};
      \addplot[black,fill=orange!40!white, postaction={pattern = north east lines, pattern color=orange!60!black}] coordinates {(4, 0.9401)};
    \end{axis}
  \end{scope}

  \begin{scope}[yshift=-.45\linewidth]
    \begin{axis}[
      width=\linewidth,
      height=.5\linewidth,
      ybar,
      xmin = 0.5,
      xmax = 4.5,
      ymin=0,
      ymax=41,
      xtick=\empty,
      ytick=\empty,
      nodes near coords,
      title={$\Delta$},
      bar width= 8mm,
      axis y line =none,
      axis y line*=none,
      axis x line*=bottom,
      axis x line =bottom,
      title style={at={(0.5, -0.1)}, anchor=north},
      x axis line style=-,
      bar shift = 0,
    ]
      \addplot[black,fill=red!40!white, postaction={pattern = crosshatch dots, pattern color=red!60!black}] coordinates {(1, 3.6340)};
      \addplot[black,fill=blue!40!white, postaction={pattern = crosshatch, pattern color=blue!60!black}] coordinates {(2, 3.5931)};
      \addplot[black,fill=green!40!white, postaction={pattern = north west lines, pattern color=green!60!black}] coordinates {(3, 40.8976)};
      \addplot[black,fill=orange!40!white, postaction={pattern = north east lines  , pattern color=orange!60!black}] coordinates {(4, 40.1287)};
    \end{axis}
  \end{scope}
\end{tikzpicture}
    }
    \caption{Mean $F_{score}$ and $\Delta$ values for \proposedname, \textit{PD-T}, \textit{PD-E}, \textit{TPCDD}.}
    \label{figure:validation:results:comparative}
  \end{figure}
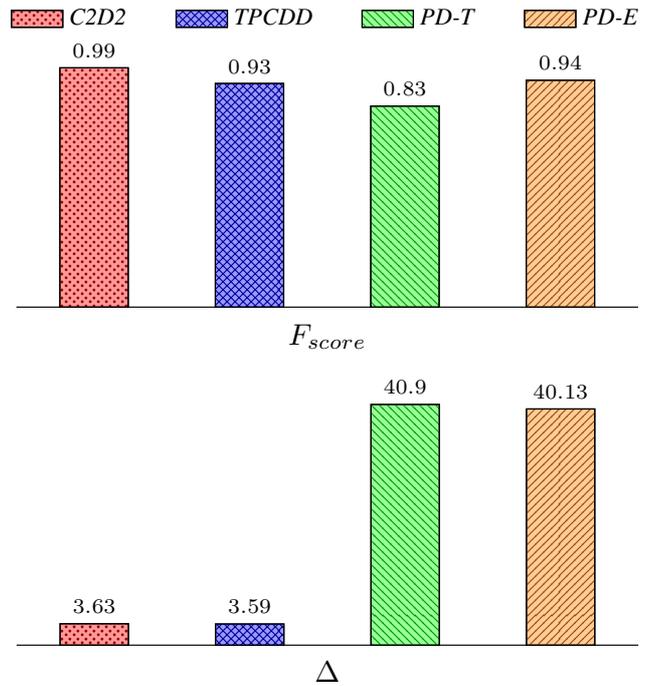

  \begin{table}[t]
    \centering
    \caption{$F_{score}$ one-vs-all statistical test results.}
    \label{table:validation:statistical-test-ranks:fscore}
\tabulinesep=2pt
\begin{tabu} to 0.45\linewidth {SX[c]}
  \textbf{Rank} & \textbf{Algorithm} \\ \toprule
  85.5221       & \textit{\proposedname}\\
 146.6985       & \textit{PD-T}             \\
 152.2132       & \textit{PD-E}             \\
 162.5662       & \textit{TPCDD}            \\ \bottomrule
\end{tabu}

  \end{table}

  \begin{table}[t]
    \centering
    \caption{$F_{score}$ post-hoc statistical test results.}
    \label{table:validation:statistical-test-posthoc:fscore}
\tabulinesep=2pt
\begin{tabu} to 0.96\linewidth {cX[c]r}
\textbf{Algorithms}              & $\mathbf{p_{value}}$   & \textbf{Result}   \\ \toprule
\proposedname~vs. \textit{PD-T}  &          < \num{1e-10} & $H_0$ is rejected \\
\proposedname~vs. \textit{PD-E}  &          < \num{1e-10} & $H_0$ is rejected \\
\proposedname~vs. \textit{TPCDD} &          < \num{1e-10} & $H_0$ is rejected \\ \bottomrule
\end{tabu}

  \end{table}

  \begin{table}[t]
    \centering
    \caption{$\Delta$ one-vs-all statistical test results.}
    \label{table:validation:statistical-test-ranks:delay}
\tabulinesep=2pt
\begin{tabu} to 0.45\linewidth {SX[c]}
  \textbf{Rank} & \textbf{Algorithm} \\ \toprule
  62.8629       & \textit{TPCDD}     \\
  63.7984       & \proposedname      \\
 185.6129       & \textit{PD-E}      \\
 185.7258       & \textit{PD-T}      \\ \bottomrule
\end{tabu}

  \end{table}

  \begin{table}[t]
    \centering
    \caption{$\Delta$ post-hoc statistical test results.}
    \label{table:validation:statistical-test-posthoc:delay}
\tabulinesep=2pt
\begin{tabu} to 0.96\linewidth {cX[c]r}
\textbf{Algorithms}              & {$\mathbf{p_{value}}$} & \textbf{Result}   \\ \toprule
\proposedname~vs. \textit{PD-T}  &          < \num{1e-10} & $H_0$ is rejected \\
\proposedname~vs. \textit{PD-E}  &          < \num{1e-10} & $H_0$ is rejected \\
\proposedname~vs. \textit{TPCDD} &           \num{0.9431} & $H_0$ is accepted \\\bottomrule
\end{tabu}

  \end{table}

  As a summary, \figurename~\ref{figure:validation:results:comparative} shows the mean $F_{score}$ and the mean $\Delta$ for each algorithm.
  As can be seen, \proposedname~outperforms every other algorithm in mean $F_{score}$, while maintaining a very low $\Delta$.
  Furthermore, $F_{score}$ results have been evaluated using a statistical test in order to confirm the quality of the proposed method.
  A \emph{non-parametric} \emph{one-vs-all} test has been executed using the STAC tool\footnote{tec.citius.usc.es/stac}\cite{DBLP:conf/fuzzIEEE/Rodriguez-FdezC15}.
  Namely, \emph{Friedman's Aligned Ranks} test with a significance level of \num{0.05} has been used.
  The test result confirms that $H_0$ is rejected, thus algorithms do not converge to the same mean, being \proposedname~statistically the best in terms of $F_{score}$.
  \tablename~\ref{table:validation:statistical-test-ranks:fscore} shows the rankings of the algorithms.
  In this case, \proposedname~clearly outscores the other algorithms.

  A post-hoc test (namely \emph{Holm} post-hoc test), a pairwise comparison, was also performed.
  Results are shown in \tablename~\ref{table:validation:statistical-test-posthoc:fscore}.
  For all the cases, $H_0$ is rejected, meaning that no algorithm is able to equal \proposedname~in $F_{score}$ values.
  Finally, tests were also conducted for $\Delta$ values.
  \proposedname~and \textit{TPCDD} are tied as the best-performing algorithms.
  \emph{Friedman's Aligned Ranks} for $\Delta$ is depicted in \tablename~\ref{table:validation:statistical-test-ranks:delay}.
  Both \proposedname~and \textit{TPCDD} obtain similar scores, thus rejecting the hypothesis that one of the algorithms outperforms the others.
  \tablename~\ref{table:validation:statistical-test-posthoc:delay} shows the post-hoc test for $\Delta$.
  As we can see in the last row, the null hypothesis $H_0$ between \proposedname and \textit{TPCDD} cannot be rejected, meaning that both algorithms converge to the same mean, which is consistent with the scores obtained by \emph{Friedman's Aligned Ranks}.

\section{Conclusions and Future Work}\label{section:conclusions}
In this paper we presented \proposedname, an approach to the \emph{offline} detection of \emph{sudden control-flow drifts} in process mining.
\proposedname~drift detection is supported by the hypotheses of conformance checking measures being suitable to detect control-flow drifts.
Specifically, we argue that fitness and precision are complementary metrics in concept drift, and while fitness is usefull to identify traces that are not supported by the model, precision looks for behaviour not present in the window of traces.
Related to this, we propose the usage of two approximations for each one of the metrics that have a low computational complexity and to detect changes in models.

Our approach has been validated against a synthetic benchmarking dataset formed by 68 logs, outperforming the best concept drift algorithms in terms of accuracy ($F_{score}$) while maintaining a minimum delay ($\Delta$).
Finally, we did a statistical test over the results of all the algorithms to confirm that the presented solution is statistically better in terms of accuracy.

As future work, we plan to study the usage of memory mechanisms in order to improve the results.
We plan also to extend the method to deal with other types of changes, as the gradual and incremental drifts, and to be executed in \emph{online} environments, where the requirements in terms of computational complexity are different.


\section*{Acknowledgements}\label{section:ack}
  The work from Víctor J. Gallego was supported by the Spanish Ministry of Education and Votational Training (grant FPU17/05138, co-funded by the European Regional Development Fund - ERDF program); the Galician Ministry of Culture, Education and Universities (accreditation 2016-2019, ED431G/08) and the European Regional Development Fund (ERDF).
  This paper was also supported by the Spanish Ministry of Economics and Competitiveness under the projects TIN2015-73566-JIN and TIN2017-84796-C2-1-R.

\bibliographystyle{IEEEtran}
\bibliography{bibliography/bibliography-short}


\begin{IEEEbiography}[{\includegraphics[width=1in,height=1.25in,clip,keepaspectratio]{example-image-a}}]
  {Víctor Gallego-Fontenla} is working towards the Ph.D degree at CiTIUS, University of Stantiago de Compostela, Spain.
  He received the BSc degree in computer science from the University of Stantiago de Compostela in 2015, and a MSc degree in artificial intelligence from the International University Menendez Pelayo (UIMP) in 2017.
  His main research interests include process mining and the analysis of changes in bussiness processes.
\end{IEEEbiography}

\begin{IEEEbiography}[{\includegraphics[width=1in,height=1.25in,clip,keepaspectratio]{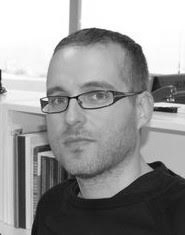}}]
  {Dr. Juan C. Vidal} received the B.Eng. degree in computer science from the University of A Coruña, A Coruña, Spain, in 2000 and his Ph.D. degree in Artificial Intelligence from the University of Santiago de Compostela (USC), Santiago de Compostela, Spain, in 2010.
  He is currently associate researcher at the Research Center in Intelligent Technology (CiTIUS), USC.
  His main research interests include linguistic data process mining, fuzzy logic, machine learning and linguistic summarization of data with fuzzy operators.
\end{IEEEbiography}

\begin{IEEEbiography}[{\includegraphics[width=1in,height=1.25in,clip,keepaspectratio]{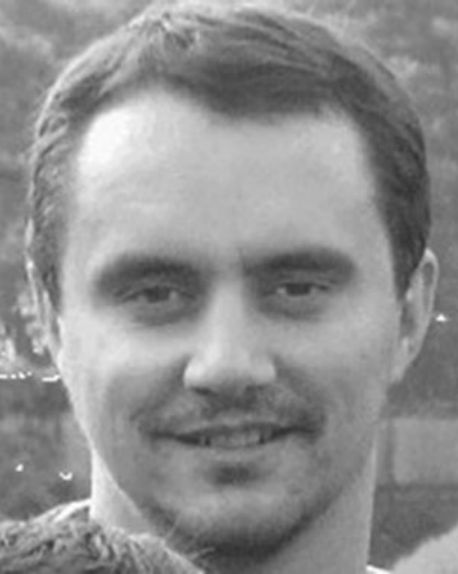}}]
  {Dr. Manuel Lama Penín} is an Associate Professor of Computer Science and Artificial Intelligence of the University of Santiago de Compostela (USC) and senior researcher of the Intelligent Technologies Center of the USC (CiTIUS).
  He is author/co-author of 140 scientific papers, 40 of them in national and international journals and books.
  He has participated in more than 30 R\&D projects and contracts, leading 20 of them.
  His research interests focus on the development of process mining techniques in massive data environments and the application of these techniques to solve real problems in areas such as Education, Health, e-commerce or Public Administrations.
\end{IEEEbiography}

\vfill

\end{document}